\documentclass[letterpaper,10pt,journal,twoside]{ieeetran} 
\IEEEoverridecommandlockouts

\pdfoutput=1

\usepackage{amsmath} 
\usepackage{amssymb}
\usepackage{amsthm} 

\usepackage{multicol}
\usepackage[bookmarks=true]{hyperref}

\usepackage{xparse} 
\usepackage{amsfonts} 
\usepackage{dsfont}
\usepackage{graphicx}
\usepackage{url}
\usepackage{booktabs}
\usepackage[para]{threeparttable}
\usepackage{multirow}
\hypersetup{colorlinks,linkcolor={blue},citecolor={green},urlcolor={blue}}  

\usepackage{xcolor}
\usepackage{framed}
\usepackage{caption}
\usepackage{varwidth}
\usepackage{subcaption}
\usepackage[utf8]{inputenc}
\usepackage{pgfplots}
\usepackage{adjustbox}
\DeclareUnicodeCharacter{2212}{−}
\usepgfplotslibrary{groupplots,dateplot}
\usetikzlibrary{patterns,shapes.arrows}
\pgfplotsset{compat=newest}
\usepackage{balance}

\usepackage{pifont}
\newcommand{\cmark}{\ding{51}}%
\newcommand{\xmark}{\ding{55}}%

\usepackage[ruled,vlined]{algorithm2e}
\SetKwInOut{Parameters}{Parameters}

\usepackage{setspace}

\usepackage{cleveref}

\usepackage{xspace}

\colorlet{shadecolor}{gray!10}
   
\newtheorem{problem}{Problem}

\NewDocumentCommand\bbm{}{ \begin{bmatrix} }
\NewDocumentCommand\ebm{}{ \end{bmatrix} }
\NewDocumentCommand\Vector{m}{ \boldsymbol{\mathbf{#1}} }
\NewDocumentCommand\Matrix{m}{ \boldsymbol{\mathbf{#1}} }

\NewDocumentCommand\Trace{m}{ \mathrm{tr}\left(#1\right) }

\NewDocumentCommand\Norm{m}{\left\Vert#1\right\Vert }

\NewDocumentCommand\ArgMin{m}{ \operatorname*{argmin}_{#1} }

\NewDocumentCommand\Set{m}{ \left\{#1\right\} }
\NewDocumentCommand\Cardinality{m}{ \left\vert#1\right\vert }
\NewDocumentCommand\Real{}{ \mathbb{R} }
\NewDocumentCommand\Natural{}{ \mathbb{N} }
\NewDocumentCommand\NonNegativeReal{}{ \Real_+ }

\NewDocumentCommand\Sym{}{ \mathbb{S} }
\NewDocumentCommand\SymmetricMatrices{m}{\Sym^{#1}}
\NewDocumentCommand\PSDMatrices{m}{\Sym^{#1}_+}

\NewDocumentCommand\HyperTorus{m}{\mathcal{T}^{#1}}

\NewDocumentCommand\LieGroupO{m}{ \mathrm{O}(#1) }

\NewDocumentCommand\AbsoluteValue{m}{ \Cardinality{m} }

\NewDocumentCommand\Identity{}{ \Matrix{I} }

\NewDocumentCommand\T{}{\mathsf{T}}

\NewDocumentCommand\diag{m}{\text{diag}\left(#1\right)}

\NewDocumentCommand\RankFunction{m}{ \mathrm{rank}(#1) }
\NewDocumentCommand\LinearMap{}{ \mathcal{A} }
\NewDocumentCommand\LinearInequalityMap{}{ \mathcal{B} }

\NewDocumentCommand\Indices{m}{[#1]}
\NewDocumentCommand\Ours{}{\texttt{CIDGIK}\xspace}
\NewDocumentCommand\SLSQP{}{\texttt{SLSQP}\xspace}
\NewDocumentCommand\IPOPT{}{\texttt{IPOPT}\xspace}

\NewDocumentCommand\Dim{}{ d }
\NewDocumentCommand\Configuration{}{ \mathcal{C} }
\NewDocumentCommand\WorkSpace{}{ \mathcal{W} }

\NewDocumentCommand\Anchor{}{ \Vector{w} }
\NewDocumentCommand\AnchorMatrix{}{ \Matrix{W} }
\NewDocumentCommand\Distance{}{ \ell }

\NewDocumentCommand\Obstacles{}{ \mathcal{O} }
\NewDocumentCommand\GoalAnchor{}{ \Anchor^{\mathrm{g}} }

\NewDocumentCommand\Graph{}{ \mathcal{G} }
\NewDocumentCommand\Vertices{}{ \mathcal{V} }
\NewDocumentCommand\Edges{}{ \mathcal{E} }
\NewDocumentCommand\EqualityEdges{}{ \Edges_{\textrm{eq}} }
\NewDocumentCommand\IncidenceMatrix{m}{ \Matrix{B}(#1) }
\NewDocumentCommand\JointVertices{}{ \Vertices_{\mathrm{j}} }
\NewDocumentCommand\JointEdges{}{ \Edges_{\mathrm{l}} }
\NewDocumentCommand\AnchorEdges{}{ \Edges_{\mathrm{w}} }
\NewDocumentCommand\PlaneVertices{}{ \Vertices_{\mathrm{p}} }

\newlength{\minuslength}
\begin{document}

\title{Convex Iteration for Distance-Geometric Inverse Kinematics}
\author{Matthew Giamou,$^{a,\dagger}$ Filip Mari\'c,$^{a,b,\dagger}$ David M. Rosen,$^c$ Valentin Peretroukhin,$^d$\\ Nicholas Roy,$^d$ Ivan Petrovi\'c,$^b$ and Jonathan Kelly$^a$
\thanks{Manuscript received: September 3, 2021; Revised December 5, 2021; Accepted January 5, 2021. This paper was recommended for publication by Editor L. Pallottino upon evaluation of the Associate Editor and Reviewers' comments. This work was supported in part by the Natural Sciences and Engineering Research Council of Canada and by a Dean's Catalyst Professorship from the University of Toronto.}
\thanks{${^\dagger}$Denotes equal contribution.}
\thanks{${^a}$Institute for Aerospace Studies, University of Toronto, Canada. {\tt\footnotesize <firstname>.<lastname>@robotics.utias.utoronto.ca}}
\thanks{${^b}$Faculty of Electrical Engineering and Computing, University of Zagreb.}
\thanks{${^c}$Laboratory for Information and Decision Systems, Massachusetts Institute of Technology.}
\thanks{${^d}$Computer Science and Artificial Intelligence Laboratory, Massachusetts Institute of Technology.}
\thanks{Digital Object Identifier (DOI): see top of this page.} }

\maketitle
\markboth{IEEE Robotics and Automation Letters. Preprint Version. Accepted January, 2022}
{Giamou \MakeLowercase{\textit{et al.}}: Convex Iteration for Inverse Kinematics} 


\begin{abstract}
Inverse kinematics (IK) is the problem of finding robot joint configurations that satisfy constraints on the position or pose of one or more end-effectors.
For robots with redundant degrees of freedom, there is often an infinite, nonconvex set of solutions.
The IK problem is further complicated when collision avoidance constraints are imposed by obstacles in the workspace.
In general, closed-form expressions yielding feasible configurations do not exist, motivating the use of numerical solution methods.
However, these approaches rely on local optimization of nonconvex problems, often requiring an accurate initialization or numerous re-initializations to converge to a valid solution.
In this work, we first formulate inverse kinematics with complex workspace constraints as a convex feasibility problem whose low-rank feasible points provide exact IK solutions. 
We then present \texttt{CIDGIK} (Convex Iteration for Distance-Geometric Inverse Kinematics), an algorithm that solves this feasibility problem with a sequence of semidefinite programs whose objectives are designed to encourage low-rank minimizers. 
Our problem formulation elegantly unifies the configuration space and workspace constraints of a robot: intrinsic robot geometry and obstacle avoidance are both expressed as simple linear matrix equations and inequalities.
Our experimental results for a variety of popular manipulator models demonstrate faster and more accurate convergence than a conventional nonlinear optimization-based approach, especially in environments with many obstacles.

\end{abstract}


\section{Introduction}
\urlstyle{tt}
\IEEEPARstart{S}{olving} inverse kinematics (IK) is an essential step for motion planning with articulated robots.
However, an efficient algorithm with a high success rate for robots with redundant degrees of freedom in obstacle-laden workspaces remains elusive.
Solving this problem would help enable fast and reliable autonomous mobility for manipulators, humanoid robots, and other articulated mechanisms.
Optimization-based approaches typically use the joint angles of the robot as decision variables.
While low-dimensional, this parameterization leads to nonconvex cost and constraint functions involving the product of multiple trigonometric functions of the joint angles. 
This nonconvexity makes finding global minima challenging for numerical solvers.

\begin{figure}
\centering 
\includegraphics[width=\columnwidth]{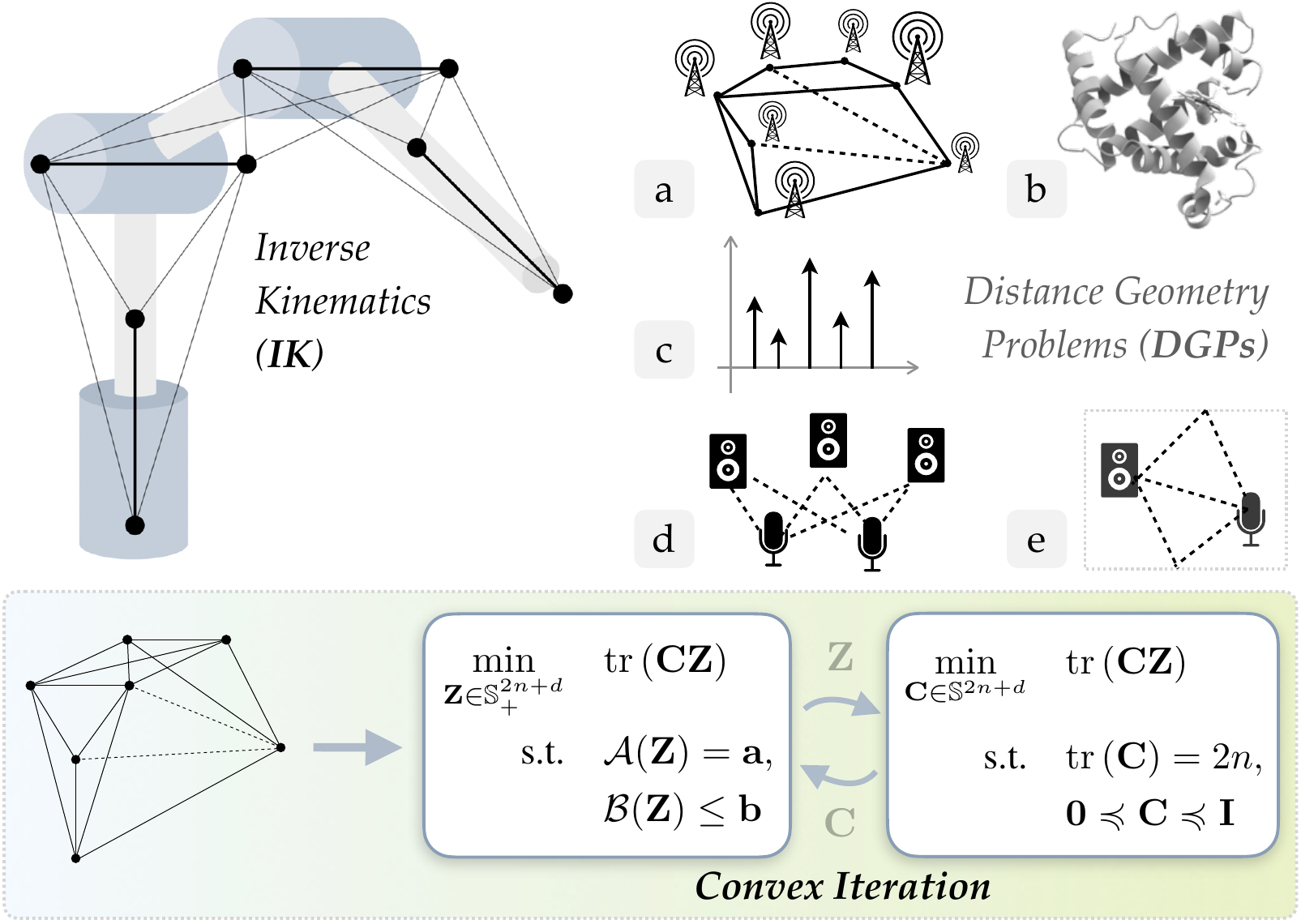}
\caption{We apply a distance-geometric formulation of IK to develop a fast and accurate IK solver based on low-rank convex optimization. Our formulation connects IK to the rich literature on convex relaxations for other distance geometry problems such as: a) sensor network localization, b) molecular conformation, c) sparse phase retrieval, d) microphone calibration, and e) indoor acoustic localization.}
\vspace{-0.7cm}
\label{fig:main_figure}
\end{figure}

Recently, a number of IK techniques have utilized alternative parameterizations based on the distances between points fixed to the robot~\cite{porta_inverse_2005, Le_Naour_2019, blanchini_convex_2017}. 
This \emph{distance-geometric} view of IK increases the number of parameters needed, but elegantly describes the workspace of a robot with simple pairwise distance constraints. 
In this paper, we connect IK to the classical distance geometry problem (DGP), which has found application in many domains, including those in \Cref{fig:main_figure} and \Cref{tab:dgps}.
Noting similarities between the IK problem and DGPs such as sensor network localization (SNL), our main contribution is to leverage the mature literature on semidefinite programming (SDP) relaxations for DGPs to develop \Ours (Convex Iteration for Distance-Geometric Inverse Kinematics), a novel IK solver.
\Ours applies a fast, minimal-rank-promoting algorithm~\cite{dattorro_convex_2005} to an SDP relaxation of our formulation of IK, encompasses a wide variety of redundant robot models, and naturally incorporates spherical obstacles and planar workspace constraints.
We provide a free Python implementation\footnote{Code: \url{https://github.com/utiasSTARS/graphIK}.} of \Ours along with experiments demonstrating its superior success rate and speed when compared with a standard nonlinear approach to solving the conventional angular formulation of IK.

\section{Related Work}
\label{sec:related_work}
In this section we briefly review the state of the art in IK, distance geometry, and semidefinite optimization.
We emphasize the intersection of these three fields, and contrast the properties of a variety of recent IK solvers.

\subsection{Inverse Kinematics}
Simple closed-form solutions can be derived for common manipulator robots that do not possess redundant degrees of freedom~\cite{spong2020robot}.
For example, kinematic chains with six revolute degrees of freedom (DOF) possess at most 16 solutions for IK problems with pose goals~\cite{manocha_efficient_1994}.
This paper deals with the more challenging case of IK with collision avoidance for redundant manipulators with one or more end-effectors.
First-order methods for IK \cite{buss_introduction_2004} are a mature and popular choice for a broad class of problems in robotics and graphics~\cite{aristidouInverseKinematicsTechniques2018}.
TRAC-IK \cite{beeson2015trac} is a fast and freely-available software implementation of IK \emph{without} collision avoidance for redundant manipulators that concurrently runs an inverse Jacobian algorithm and a sequential quadratic programming algorithm. 
A recent mixed-integer programming approach \cite{dai_global_2019} provides approximate solutions with global guarantees for a general formulation of IK with joint limits, multiple end-effectors, and expressive workspace constraints (e.g., specifying free space with polyhedra).
In spite of its high success rate on challenging problems, the approximate nature and long runtime (on the order of 15 seconds for an 18-DOF quadruped) prohibit the use of \cite{dai_global_2019} in realtime applications.

\subsection{Distance Geometry}
Euclidean distance geometry is concerned with estimating the positions of points given a subset of their pairwise distances~\cite{libertiEuclideanDistanceGeometry2014, dattorro_convex_2005}.
Applications, some of which are depicted in \Cref{fig:main_figure}, include problems in computational chemistry, machine learning, and signal processing~\cite{dokmanic_euclidean_2015}.
A distance-geometric formulation of IK is presented in \cite{porta_inverse_2005} for a family of common revolute robot manipulators. 
This formulation is solved with the complete but computationally expensive (and therefore inappropriate for real-time robotics applications) branch-and-prune solver introduced in \cite{porta_branch-and-prune_2005}. 
Other approaches to IK that incorporate distance constraints on point variables include the unconstrained quartic optimization approach to IK for graphics found in \cite{Le_Naour_2019} and the convex relaxation for simple robotic chains in \cite{blanchini_inverse_2015, blanchini_convex_2017}.
In this work, we apply methods and analysis based on SDP relaxations for SNL~\cite{so_theory_2007, dattorro_convex_2005} to a formulation inspired by \cite{porta_inverse_2005}.
We characterize the similarities and differences between IK and other DGPs in \Cref{tab:dgps} and \Cref{sec:semidefinite}. 

\begin{table}[ht]
\small
\centering
\begin{tabular}{l | cccc}
& \multicolumn{3}{c}{Distance Measurements} & Under \\
Application & \textit{missing} & \textit{noisy} & \textit{free}$^{\dagger}$ & Dtmnd. \\
\midrule
Wireless sensor networks & \cmark & \cmark & \xmark & \xmark \\
Molecular conformation & \cmark & \cmark & \xmark & \cmark/\xmark \\
Sparse phase retrieval & \xmark& \cmark & \cmark & \xmark \\
Microphone calibration & \cmark & \cmark & \xmark & \xmark \\
Indoor acoustic localization & \xmark & \cmark & \cmark & \xmark \\
\midrule
Inverse kinematics & \cmark & \xmark & \xmark & \cmark \\
\end{tabular}
\caption{Properties of different types of DGPs (extended from \cite{dokmanic_euclidean_2015}). IK is unique in that it deals with ``noiseless" (i.e., exact) geometric constraints \emph{and} is underdetermined because of the existence of multiple valid solutions. $^{\dagger}$Free (or \textit{unlabeled}) measurements are those which lack a known association with two points.}
\label{tab:dgps}
\end{table}

\subsection{Semidefinite Programming}

Many parameter estimation problems in statistics and engineering can be expressed as quadratically-constrained quadratic programs (QCQPs) \cite{cifuentes_local_2020, so_theory_2007}.
While difficult to solve in the worst case, primal SDP relaxations of QCQPs~\cite{boyd_semidefinite_1997}, which we employ in \Cref{sec:semidefinite}, can often be efficiently solved by interior-point methods \cite{so_theory_2007}.
Remarkably, in many cases, this relaxation is provably \textit{tight} under the assumption of low noise \cite{cifuentes_local_2020} and one can recover the global optimum by solving the relaxed problem. 
Furthermore, many of these problems exhibit structure (e.g., chordal sparsity) that can be exploited for improved performance over generic SDP solvers on large-scale problem instances~\cite{majumdar_survey_2019}.
In \cite{trutman_globally_2020}, Lasserre's hierarchy of SDP relaxations is applied to IK for 7-DOF revolute manipulators, but solutions are prohibitively slow for real-time applications. 

Our approach extends the distance-geometric description of the kinematics of a revolute robot in \cite{maric_riemannian_2021} to a form very similar to the SNL problem.
Building on our work with planar and spherical chains in \cite{maricInverseKinematicsSerial2020}, we analyze and solve this formulation using the rich literature on SDP relaxations for SNL \cite{so_theory_2007, nie_sum_2009, dingSensorNetworkLocalization2010}, while noting the properties in \Cref{tab:dgps} that make IK distinct from SNL.
Our approach is similar to the work presented in \cite{yenamandra_convex_2019}, where the authors apply a different convex relaxation to a formulation of IK for arbitrary joint angle-limited tree-like revolute models. 
However, they do not incorporate obstacle avoidance, report runtimes in excess of 10 seconds, and their relaxation only provides coarse initializations for a local solver. 

\section{Problem Formulation} \label{sec:problem_formulation}
We begin with a summary of our notation, followed by a detailed description of our ``distance-geometric" formulation of robot kinematics.

\subsection{Notation}
%
Boldface lower and upper case letters (e.g., $\Vector{x}$ and $\Matrix{P}$) represent vectors and matrices respectively. 
The bracketed superscript in $\Matrix{A}^{(i)}$ indicates the $i$th column of the matrix $\Matrix{A}$. 
We write $\Identity_{n}$ (or $\Identity$ when clear from context) for the $n \times n$ identity matrix.
%
%
The space of $n\times n$ symmetric and symmetric positive semidefinite (PSD) matrices are denoted $\SymmetricMatrices{n}$ and $\PSDMatrices{n}$, respectively, and we also write $\Matrix{A} \succeq \Matrix{B}$ ($\Matrix{A} \succ \Matrix{B}$) to indicate that $\Matrix{A} - \Matrix{B}$ is PSD (positive definite).
We denote the set of indices $\Set{1, \ldots, n}$ as $\Indices{n}$ for any $n \in \Natural$.
Finally, $\Norm{\cdot}$ always represents the Euclidean norm.

\begin{figure}
  \centering
	\begin{subfigure}{0.49\columnwidth}
    \centering
    \includegraphics[width=\textwidth]{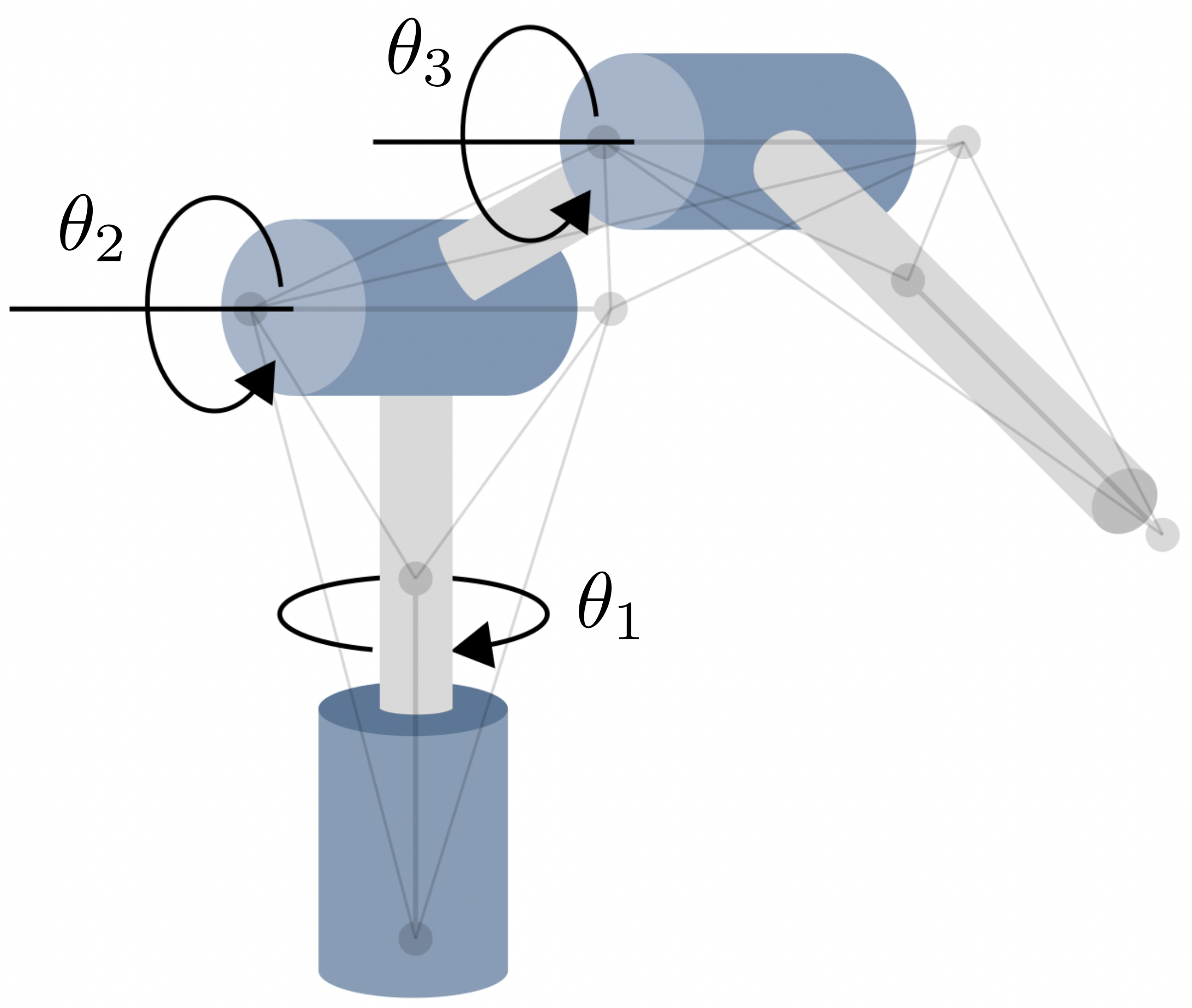}
    \caption{}\label{fig:revolute_chain_a}
  \end{subfigure}
	\begin{subfigure}{0.49\columnwidth}
    \centering
    \includegraphics[width=\textwidth]{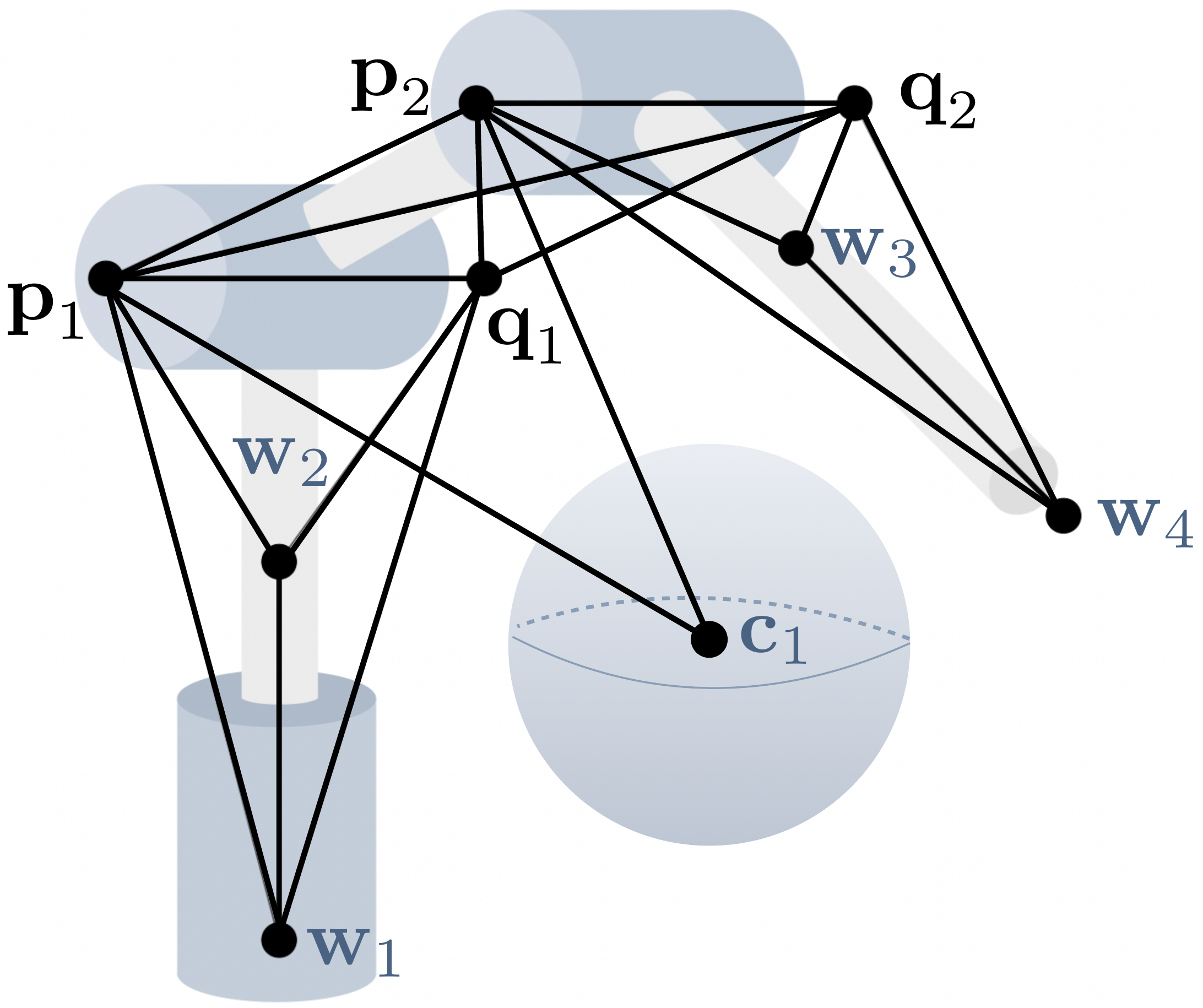}
    \caption{}\label{fig:revolute_obstacles}
  \end{subfigure}
  \caption{Visualization of a 3-DOF revolute manipulator: a) Actuated joints and links, overlaid with the graph of known distances. b) Depiction of our distance-geometric formulation with a spherical obstacle (cf. \Cref{sec:kinematic_model}).}\label{fig:revolute_chain_big}
  \vspace{-5mm}
\end{figure}

\subsection{Kinematic Model}
\label{sec:kinematic_model}
Our proposed kinematic model eschews joint angle variables in favour of points embedded in $\Real^\Dim$ \cite{porta_inverse_2005}, where $\Dim \in \{2, 3\}$ for physically realizable revolute robots.
These points are strategically fixed relative to an $N$-DOF robot's articulated joints such that their positions fully describe the underlying angular configuration $\Vector{\Theta} \in \Configuration \subseteq \HyperTorus{N}$, where $\HyperTorus{N}$ is the $N$-dimensional torus. 
IK is typically expressed as finding, for a given goal $\GoalAnchor$, joint angles $\Vector{\Theta} \in \Configuration$ that satisfy the system of equations
\begin{equation} \label{eq:forward_kinematics}
	F(\Vector{\Theta}) = \GoalAnchor \in \WorkSpace,
\end{equation}
where $F: \Configuration \rightarrow \WorkSpace$ is the trigonometric \emph{forward} kinematics function that maps joint angles in the configuration space $\Configuration$ to end-effector positions or poses in the workspace $\WorkSpace$.

When a closed form solution $\Vector{\Theta} = F^{-1}(\GoalAnchor)$ is unavailable, numerical methods are typically used to solve optimization-based formulations of IK \cite{erleben_solving_2019}:
\begin{problem}[Inverse Kinematics] \label{prob:inverse_kinematics}
Given a robot's forward kinematics $F: \Configuration \rightarrow \WorkSpace$ and desired end-effector position(s) or pose(s) $\Anchor \in \WorkSpace$, find joint angles $\Vector{\Theta}$ that solve
\begin{equation}
\min_{\Vector{\Theta} \in \Configuration} \quad \Norm{F(\Vector{\Theta}) - \GoalAnchor}^2.
\end{equation}
\end{problem}
\noindent Our formulation, which is illustrated for a simple 3-DOF manipulator in \Cref{fig:revolute_chain_big}, specifies the goal poses or positions of a robot's end-effector(s) and base using $m \geq 2$ points or \emph{anchors} $\Anchor_k \in \Real^\Dim$ for $k \in \Indices{m}$.
In contrast with the traditional angular formulation of \Cref{prob:inverse_kinematics} displayed in \Cref{fig:revolute_chain_a}, the state space of our formulation is comprised of points rigidly fixed to $n$ ``unanchored" joints.\footnote{A joint is \emph{anchored} if its axis of rotation is invariant to changes in $\Matrix{\Theta}$.} 
For the example in \Cref{fig:revolute_chain_big}, $n=2$ and the unanchored joints are those actuated by $\theta_2$ and $\theta_3$, while the joint actuated by $\theta_1$ is anchored.
To enforce rigid link lengths that are invariant to the robot's angular configuration $\Matrix{\Theta}$, each unanchored joint $i \in \Indices{n}$ is assigned a pair of points $\Vector{p}_i, \Vector{q}_i$ positioned along its axis of rotation as in \Cref{fig:revolute_obstacles} such that $\Norm{\Vector{p}_i - \Vector{q}_i} = 1 \enspace \forall \ \Matrix{\Theta}$. 
Since the relative pose between consecutive joints $i$ and $j$ is fixed by a rigid link, each pair of joints is described by the six pairwise distance constraints between $\Vector{p}_i$, $\Vector{q}_i$, $\Vector{p}_j$, and $\Vector{q}_j$ shown as lines in \Cref{fig:revolute_obstacles}. 
Additionally, in our formulation, a robot's fixed base is treated in a manner identical to end-effectors (e.g., $\Anchor_1$ and $\Anchor_2$ in \Cref{fig:revolute_obstacles}).

End-effector position and pose targets are enforced via the fixed distance constraints between unanchored joints and neighbouring anchors (e.g., $\Norm{\Vector{p}_2 - \Anchor_3}$ and $\Norm{\Vector{q}_1 - \Anchor_1}$ in \Cref{fig:revolute_obstacles}).
A problem instance is defined by an assignment of all anchors $\AnchorMatrix = [\Anchor_1 \ \cdots \ \Anchor_m] \in \Real^{\Dim \times m}$.
For example, an end-effector's position can be constrained by specifying a single anchor position on the tip of the end-effector (e.g., $\Anchor_4$ in \Cref{fig:revolute_obstacles}).
The ``direction" (i.e., orientation without yaw), of this end-effector can be constrained by specifying the position of an additional distinct anchor (e.g., $\Anchor_3$ in \Cref{fig:revolute_obstacles}) along the desired direction~\cite{maric_riemannian_2021}.

In order to form a graph describing our kinematic model, let $\JointVertices = \Indices{2n}$ and $\Vertices_{\mathrm{w}} = \Indices{2n+m} \setminus \Indices{2n}$ be index sets for vertices representing variable points ($\Vector{p}_i$ and $\Vector{q}_i$) and anchors ($\Anchor_j$) respectively, and let $\Vertices = \JointVertices \cup \Vertices_{\mathrm{w}} = \Indices{2n+m}$.
Similarly, the edge sets describing fixed distance constraints between variable joints and between variable joints and anchors are $\JointEdges \subseteq \JointVertices \times \JointVertices$ and $\AnchorEdges \subseteq \JointVertices \times \Vertices_{\mathrm{w}}$, respectively. 
We represent the equality constraints with a weighted directed acyclic graph $\Graph = (\Vertices, \EqualityEdges, \Distance)$, where $\Distance: \EqualityEdges \rightarrow \NonNegativeReal$ encodes the fixed distances:
\begin{equation}
	\Distance: (i, j) \mapsto \Norm{\Vector{x}_i - \Vector{x}_j}, 
\end{equation}
where $\Vector{x}_i$ and $\Vector{x}_j$ can refer to either variable joints or fixed anchors, and $\EqualityEdges = \JointEdges \cup \AnchorEdges$.
The incidence matrix $\IncidenceMatrix{\EqualityEdges} \in \Real^{\Cardinality{\Vertices} \times \Cardinality{\EqualityEdges}}$ can be used to compactly summarize the squared distance constraints as
\begin{equation} \label{eq:distance_constraints}
	\diag{\IncidenceMatrix{\EqualityEdges}^\T \Matrix{P}^\T \Matrix{P} \IncidenceMatrix{\EqualityEdges}} = \Vector{\Distance},
\end{equation}
where
\begin{align} \label{eq:distance_constraints_explained}
	\Matrix{P} &= [\Matrix{X} \ \AnchorMatrix] \in \Real^{d \times (2n + m)}, \notag \\
	\Matrix{X} &= [\Vector{p}_1 \ \Vector{q}_1 \ \cdots \ \Vector{p}_n \ \Vector{q}_n] \in \Real^{d \times 2n}, \\
	\Vector{\Distance}_e &= \Distance(e)^2 \enspace \forall e \in \EqualityEdges. \notag
\end{align}
This formulation is equivalent to the one used in \cite{so_theory_2007} for SNL, and is described in detail in our supplementary material~\cite{giamou_supplementary_2022}.

\subsection{Workspace Constraints}
Consider a workspace $\WorkSpace$ with obstacles or other regions that our robot is forbidden from occupying.
We model these constraints with a finite set of spheres $\Obstacles$ whose union is chosen to cover the restricted regions:
\begin{equation} \label{eq:spherical_obstacle}
	\Norm{\Vector{x}_i - \Vector{c}_j}^2 \geq l^2_j \enspace \forall i \in \JointVertices, \enspace \forall j \in \Obstacles,
\end{equation}
where $\Vector{c}_j \in \Real^\Dim$ is the centre and $l_j > 0$ is the radius of sphere $j \in \Obstacles$.
This ``union of balls" environment representation has been used in previous work on robot motion planning~\cite{varava_caging_2020} and can approximate complex obstacles up to arbitrary precision with a large number of spheres.
For specifying a spherical region of free space \emph{within} which some subset of the joints must lie (e.g., the obstacle-free regions computed in \cite{akin_computing_2015}), the inequality in \Cref{eq:spherical_obstacle} is simply reversed. 

Additionally, constraining a point $\Vector{x} \in \Real^\Dim$ on our robot to lie in (on one side of) a plane can be simply encoded as a linear equality (inequality) constraint with the plane's normal $\Vector{n}$ and its minimum Euclidean distance $c$ from the origin:
\begin{equation} \label{eq:planar_constraint}
	\Vector{x}^\T \Vector{n} \leqq c.
\end{equation}
One useful application of planar constraints is in restricting a legged robot's feet to lie on (or above) the floor.
Finally, self-collisions and workspace constraints on points lying \emph{between} joints (i.e., on long links) can be easily introduced by adding simple auxiliary variables to $\Matrix{X}$ and their corresponding constraints as described in \cite{giamou_supplementary_2022}.

\subsection{QCQP Formulation}
We can use the quadratic expressions of Equations (\ref{eq:distance_constraints}) and (\ref{eq:spherical_obstacle}) along with the linear constraints of \Cref{eq:planar_constraint} to describe the solutions to our IK problem as the following quadratic feasibility program:
\begin{problem}[Feasibility QCQP] \label{prob:feasibility_qcqp}
\begin{align}
\mathrm{find} &\quad \Matrix{X}  \in \Real^{d \times 2n} \notag \\
\text{\emph{s.t.}} &\quad \Matrix{P} = [\Matrix{X} \ \AnchorMatrix], \notag \\
&\quad \text{\emph{diag}}(\IncidenceMatrix{\EqualityEdges}^\T \Matrix{P}^\T \Matrix{P}\IncidenceMatrix{\EqualityEdges})) = \Vector{\Distance}, \\
&\quad \Norm{\Vector{x}_i - \Vector{c}_j}^2 \geq l^2_j \enspace \forall i \in \JointVertices, \enspace \forall j \in \Obstacles, \notag \\
&\quad \Vector{x}_i^\T \Vector{n}_i = c_i \enspace \forall i \in \PlaneVertices \subset \JointVertices \notag,
\end{align}
where $\EqualityEdges$, $\Vector{\Distance}$, and $\Matrix{W}$ are problem parameters defined by Equations (\ref{eq:distance_constraints}) and (\ref{eq:distance_constraints_explained}), $\Obstacles$ is the set of spherical obstacle constraints of the form in \Cref{eq:spherical_obstacle}, and $\PlaneVertices$ indexes the nodes confined to lie in some plane described by $\Vector{n}_i$ and $c_i$ in \Cref{eq:planar_constraint}.
\end{problem}
\noindent \emph{Remark}: When solutions exist, the feasible set of \Cref{prob:feasibility_qcqp} is nonconvex and therefore challenging to characterize. 
In fact, the analogous formulation of \Cref{prob:feasibility_qcqp} for SNL with exact distance measurements, no obstacles, and known dimension $\Dim$ is strongly NP-hard \cite{krislock_explicit_2010, aspnes_computational_2004}.

While quite general, our formulation does have a few limitations: it is only capable of describing the kinematics of an articulated robot with neighbouring joint axes that are coplanar (i.e., parallel or intersecting),\footnote{Many commercial manipulators (e.g., those in \Cref{fig:urdf_fig}), satisfy this requirement, and we discuss potential workarounds in~\cite{giamou_supplementary_2022}.} it cannot enforce arbitrary joint angle limits, and it only supports spherical and planar workspace constraints~\cite{maric_riemannian_2021}.

\section{Semidefinite Relaxations} \label{sec:semidefinite}
We turn to semidefinite programming (SDP) relaxations as a means of efficiently computing solutions to \Cref{prob:feasibility_qcqp}.
Introducing the lifted matrix variable
\begin{equation} \label{eq:lifted_Z_variable}
	\Matrix{Z}(\Matrix{X}) \triangleq [\Matrix{X} \enspace \Identity_\Dim]^\T [\Matrix{X} \enspace \Identity_\Dim] = 
	\begin{bmatrix}
	  \Matrix{X}^\T \Matrix{X} & \Matrix{X}^\T \\
	  \Matrix{X} & \Identity_\Dim	
	\end{bmatrix}
\in \PSDMatrices{2n+\Dim} 
\end{equation}
permits us to rewrite the quadratic constraints of \Cref{prob:feasibility_qcqp} as \emph{linear} functions of $\Matrix{Z}(\Matrix{X})$.
Since $\Matrix{Z}(\Matrix{X})$ is an outer product of matrices with rank of at most $d$ (the dimension of the space in which the robot operates), we know that $\RankFunction{\Matrix{Z}} \leq d$.  
Replacing $\Matrix{Z}(\Matrix{X})$ with a \emph{generic} PSD matrix $\Matrix{Z} \succeq \Matrix{0}$ produces the following semidefinite relaxation:
\begin{problem}[SDP Relaxation of \Cref{prob:feasibility_qcqp}] \label{prob:feasibility_sdp}
\begin{align}
\mathrm{find} &\quad \Matrix{Z} \in \PSDMatrices{2n+\Dim} \notag\\
\text{\emph{s.t.}} &\quad \LinearMap(\Matrix{Z}) = \Vector{a}, \\
&\quad \LinearInequalityMap(\Matrix{Z}) \leq \Vector{b}, \notag
\end{align}
where $\LinearMap: \SymmetricMatrices{2n+\Dim} \rightarrow \Real^{m+\Dim^2 + \Cardinality{\PlaneVertices}}$ and $\Vector{a} \in \Real^{m+\Dim^2+\Cardinality{\PlaneVertices}}$ encode the linear equations that enforce the constraints in  \Cref{eq:distance_constraints} and \Cref{eq:planar_constraint} after applying the substitution in \Cref{eq:lifted_Z_variable}, and the linear map $\LinearInequalityMap: \SymmetricMatrices{2n+\Dim} \rightarrow \Real^{\Cardinality{\Obstacles}}$ and vector $\Vector{b} \in \Real^{\Cardinality{\Obstacles}}$ enforce the inequalities in \Cref{eq:spherical_obstacle}~\cite{giamou_supplementary_2022}.
\end{problem}	
\noindent \Cref{prob:feasibility_sdp} is now a convex feasibility problem, which can be efficiently solved by numerous interior-point methods \cite{boyd2004convex}.
Unfortunately, solutions to \Cref{prob:feasibility_sdp} are not limited to the rank-$\Dim$ solutions originally sought in \Cref{prob:feasibility_qcqp}.
In fact, Nie \cite{nie_sum_2009} and So and Ye \cite{so_theory_2007} point out that when there are multiple possible solutions, interior-point SDP solvers return a max-rank solution. 
For the case of SNL problems with exact measurements and no inequalities, So and Ye \cite{so_theory_2007} use rigidity theory to prove that the existence of a unique solution in dimension $\Dim$ to an instance of \Cref{prob:feasibility_qcqp} is a sufficient condition for its corresponding SDP relaxation (\Cref{prob:feasibility_sdp}) to yield a rank-$\Dim$ or lower solution.
%
%
Unfortunately, even though the lengths in our kinematic model are in fact exact ``measurements", we are particularly interested in \emph{redundant} kinematic models, which, by definition, admit multiple solutions.
Thus, we turn our attention to methods for finding low-rank solutions to \Cref{prob:feasibility_sdp}.

\subsection{Rank Minimization} \label{sec:rank_minimization}
Ideally, we could augment \Cref{prob:feasibility_sdp} with $\RankFunction{\Matrix{Z}}$ as its cost function to find the lowest-rank solution possible.
However, the rank of a matrix is nonconvex and therefore difficult to globally minimize, even over the convex feasible set of \Cref{prob:feasibility_sdp}.
Thus, we minimize convex (linear) heuristic cost functions that encourage low rank solutions:
\begin{problem}[\Cref{prob:feasibility_sdp} with a Linear Cost] \label{prob:rank_min_sdp}
Find the symmetric PSD matrix $\Matrix{Z}$ that solves
\begin{align}
\min_{\Matrix{Z} \in \PSDMatrices{2n+\Dim}} &\quad \Trace{\Matrix{C}\Matrix{Z}} \notag\\
\text{\emph{s.t.}} &\quad \LinearMap(\Matrix{Z}) = \Vector{a}, \\
&\quad \LinearInequalityMap(\Matrix{Z}) \leq \Vector{b}, \notag
\end{align}
where $\Matrix{C} \in \SymmetricMatrices{2n + \Dim}$.
\end{problem}
\noindent When $\Matrix{C} = \Identity$, the cost function is the nuclear norm of $\Matrix{Z}$, which is the \emph{convex envelope} of $\RankFunction{\Matrix{Z}}$~\cite{dattorro_convex_2005}.
The nuclear norm heuristic has been successfully applied to a variety of linear inverse problems with matrix variables, and is even guaranteed to produce the minimum rank solution when certain conditions are met~\cite{recht_guaranteed_2010, chandrasekaran_convex_2012}.
However, our experiments demonstrate that the nuclear norm heuristic is unable to yield rank-$\Dim$ solutions to \Cref{prob:feasibility_sdp} for a simple 6-DOF manipulator~\cite{giamou_supplementary_2022}. 

Consider the following surrogate for the ``excess rank" (i.e., $\RankFunction{\Matrix{Z}} - \Dim$) of $\Matrix{Z} \in \PSDMatrices{2n+\Dim}$:
\begin{equation}
	h(\Matrix{Z}) = \sum_{i=\Dim +1}^{2n+\Dim} \lambda_{i}(\Matrix{Z}),
\end{equation}
where $\lambda_{i}(\Matrix{Z})$ is the $i$th largest eigenvalue of $\Matrix{Z}$.
Since $\Matrix{Z}$ has nonnegative eigenvalues, $h(\Matrix{Z}) = 0$ implies that $\RankFunction{\Matrix{Z}} \leq \Dim$.
Computing $h(\Matrix{Z})$ is equivalent to solving a particular SDP~\cite{dattorro_convex_2005}:
\begin{problem}[Sum of $2n$ Smallest Eigenvalues~\cite{dattorro_convex_2005}] \label{prob:fantope_sdp}
Find the symmetric PSD matrix $\Matrix{C}$ that solves
\begin{align}
\sum_{i=\Dim +1}^{2n+\Dim} \lambda_{i}(\Matrix{Z}) = \min_{\Matrix{C} \in \SymmetricMatrices{2n+\Dim}} &\quad \Trace{\Matrix{C}\Matrix{Z}} \notag \\
\text{\emph{s.t.}} &\quad \Trace{\Matrix{C}} = 2n, \\
&\quad \Matrix{0} \preccurlyeq \Matrix{C} \preccurlyeq \Identity. \notag 
\end{align}
\end{problem}
\noindent A closed-form solution to \Cref{prob:fantope_sdp} is given by \cite{dattorro_convex_2005}:
\begin{align}\label{eq:fantope_closed_form}\Matrix{C}^\star &= \Matrix{U}\Matrix{U}^\T, \\
	\Matrix{U} &= \Matrix{Q}(:,\ d+1:2n+d), \notag
\end{align}
\noindent where $\Matrix{Q} \in \LieGroupO{2n+d}$ is from the eigendecomposition $\Matrix{Z} = \Matrix{Q}\Matrix{\Lambda}\Matrix{Q}^\T$. 

\subsection{Convex Iteration}\label{sec:convex_iteration}
In \cite{dattorro_convex_2005}, the method of \emph{convex iteration} between \Cref{prob:rank_min_sdp} and \Cref{prob:fantope_sdp} is proposed. We summarize the approach in Algorithm \ref{alg:convex_iteration}.
\begin{algorithm}
\SetAlgoLined
\KwIn{\Cref{prob:rank_min_sdp} specification (i.e., $\LinearMap(\cdot), \LinearInequalityMap(\cdot), \Vector{a}, \Vector{b}$)}
\KwResult{PSD matrix $\Matrix{Z}^\star$ that solves \Cref{prob:rank_min_sdp}}
Initialize $\Matrix{C}^{\{k\}} = \Identity_{2n+\Dim}$ \\
\While{\textbf{not} \textbf{converged}}
{
 	Solve $\Matrix{Z}^{\{k\}} = \ArgMin{\Matrix{Z}}$ \Cref{prob:rank_min_sdp} with $\Matrix{C} = \Matrix{C}^{\{k\}}$ \\
 	Solve $\Matrix{C}^{\{k\}} = \ArgMin{\Matrix{C}}$ \Cref{prob:fantope_sdp} with $\Matrix{Z} = \Matrix{Z}^{\{k\}}$ using \Cref{eq:fantope_closed_form} \\
}
Return $\Matrix{Z}^\star = \Matrix{Z}^{\{k\}}$

\caption{Convex Iteration for Distance\\ Geometric IK (\Ours)}
\label{alg:convex_iteration}
\end{algorithm}
Each iteration of \Cref{prob:fantope_sdp} computes $\Matrix{C}^{\{k\}}$ corresponding to $h(\Matrix{Z})$ at the current iteration's value of $\Matrix{Z} = \Matrix{Z}^{\{k\}}$.
Since this $\Matrix{C}^{\{k\}}$ is only exact at $\Matrix{Z}^{\{k\}}$, each iteration of \Cref{prob:rank_min_sdp} can therefore be treated as minimizing an approximation of $h(\Matrix{Z})$ in the neighbourhood of $\Matrix{Z}^{\{k\}}$.
Since the closed-form solution in \Cref{eq:fantope_closed_form} is used to quickly solve \Cref{prob:fantope_sdp} in this procedure, most of \Ours's computational cost comes from solving \Cref{prob:rank_min_sdp}.
This approach has been successfully applied to noisy SNL~\cite{dattorro_convex_2005} and optimal power flow problems~\cite{wang_chordal_2018}. 

\subsection{Geometric Interpretation}\label{sec:geometric_interpretation}
Here, we motivate the convex iteration algorithm described in \Cref{sec:rank_minimization} and explain why we expect some $\Matrix{C} \in \PSDMatrices{2n+\Dim}$ to yield a low-rank solution. 
Aside from the interpretation of $\Trace{\Matrix{C}\Matrix{Z}}$ as a local approximation of the excess rank heuristic $h(\Matrix{Z})$, it is fruitful to view $\Matrix{C}$ as a \emph{direction} in the space  $\SymmetricMatrices{2n+\Dim} \supset \PSDMatrices{2n+\Dim}$.
More precisely: since $\partial\Trace{\Matrix{C}\Matrix{Z}} / \partial \Matrix{Z} = \Matrix{C}$, we are in effect \emph{designing} the objective in \Cref{prob:rank_min_sdp} so that its steepest descent direction at Z points towards \emph{low-rank} minimizers on the boundary of the feasible set.
In practice, this heuristic chooses rank-$\Dim$ matrices with high probability for typical IK problems.

\begin{figure}
\centering 
\includegraphics[width=0.94\columnwidth]{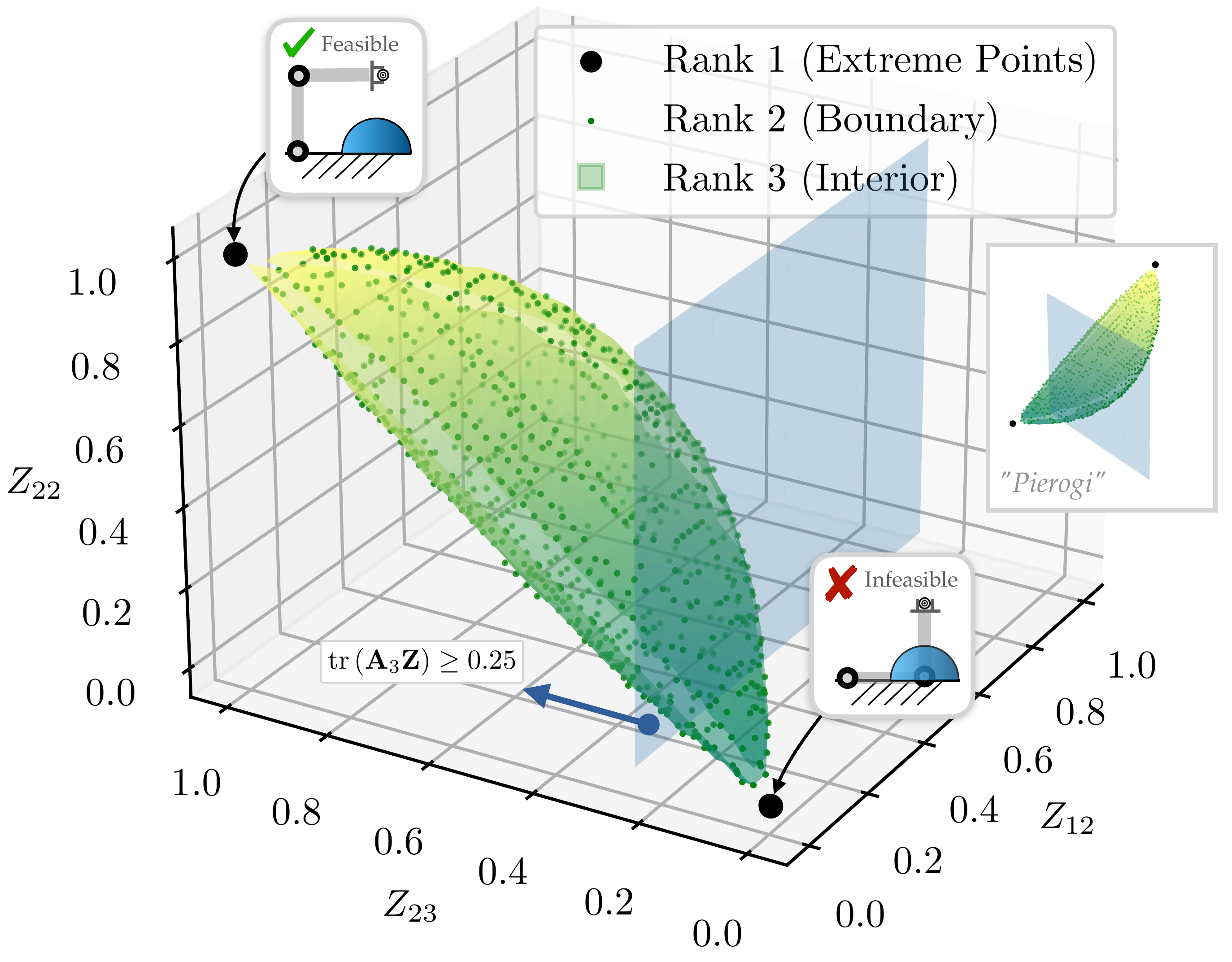}
\caption{Spectrahedron (in the shape of a \textit{pierogi}) describing the feasible set of \Cref{prob:feasibility_sdp} for the simple 2-DOF planar manipulator IK problem in \Cref{sec:geometric_interpretation}. The spectrahedron's interior (green gradient) represents the set of full-rank $3 \times 3$ PSD matrices $\Matrix{Z}$ that satisfy the constraints (linear in $\Matrix{Z}$) imposed by manipulator geometry and end-effector pose. The spectrahedron boundary (green dots) contains all degenerate $\Matrix{Z}$, and the extreme points (black dots) represent the desired rank-1 solutions. The constraint imposed by the circular obstacle is enforced in the lifted PSD representation by a half-space constraint induced by the translucent blue plane.}
\label{fig:spectrahedral_feasible_set}
\vspace{-6mm}
\end{figure}

Consider the toy problem of a 2-DOF planar manipulator rooted at the origin of the plane and with links of unit length. 
Using the formulation in \Cref{sec:problem_formulation}, we can write the IK problem of reaching a point $\Vector{w} \in \Real^2$ with the manipulator's end-effector as the following quadratic feasibility problem:
\begin{align} \label{eq:toy_qcqp}
\mathrm{find}  &\quad \Vector{x} \in \Real^2 \notag \\
\text{\emph{s.t.}}  &\quad \Norm{\Vector{x}}^2 = 1, \\
&\quad \Norm{\Vector{x} - \Vector{w}}^2 = 1, \notag \\
&\quad \Norm{\Vector{x}- \Vector{o}}^2 \geq 0.25, \notag	
\end{align}
where $\Vector{x}$ is the position of the ``elbow" joint, and $\Vector{o} = [1,\ 0]^\T$ is the position of a unit-diameter circular obstacle.
Consider the case of $\Vector{w} = [1,\ 1]^\T$: the insets of \Cref{fig:spectrahedral_feasible_set} show that of the two candidate solutions to this problem, the ``elbow down" configuration in the bottom right collides with the obstacle at $\Vector{o}$ (partially depicted as a blue semicircle). 
Homogenizing (\ref{eq:toy_qcqp}) with $s^2 = 1$ and lifting to the rank-1 matrix variable
\begin{equation}
	\Matrix{Z}(\Vector{x}) = \bbm \Vector{x} \\ s \ebm \bbm \Vector{x}^\T & s \ebm
\end{equation}
lets us apply the SDP relaxation $\Matrix{Z} \succeq \Matrix{0}$ to yield:
\begin{align} \label{eq:toy_sdp}
\mathrm{find} &\quad \Matrix{Z} \in \PSDMatrices{3} \notag \\
\text{\emph{s.t.}} &\quad \LinearMap(\Matrix{Z}) = \Vector{a},\\
&\quad \Trace{\Matrix{B}\Matrix{Z}} \leq b, \notag	
\end{align}
where the homogenization equation and the constraints of (\ref{eq:toy_qcqp}) have been replaced by their SDP equivalents~\cite{giamou_supplementary_2022}.
The intersection of the three affine equality constraints with $\Matrix{Z} \succeq \Matrix{0}$ produce the \emph{spectrahedron} in \Cref{fig:spectrahedral_feasible_set}.
Valid solutions to the unrelaxed QCQP problem are rank-1 elements of this spectrahedron.
Since the interior of this set ($\Matrix{Z} \succ \Matrix{0}$) contains full rank solutions, we know that rank-1 solutions will lie on the boundary.
Indeed, the two extreme points, denoted in black in \Cref{fig:spectrahedral_feasible_set}, represent the two valid solutions to our toy IK problem in the absence of obstacles. 
The half-space constraint induced by the blue plane illustrates the effect of the obstacle constraint in our lifted problem domain, eliminating the infeasible ``elbow down" configuration from the feasible set of (\ref{eq:toy_sdp}).
The method employed in this paper seeks to find, via convex iteration, a ``direction" $\Matrix{C}$ such that the linear cost function $\Trace{\Matrix{C}\Matrix{Z}}$ is minimized at a rank-$d$ solution.
While the solution to this toy example can be obtained analytically, its low-dimensional structure allows us to illustrate the geometric ideas motivating our approach. 
 
\section{Experiments} \label{sec:experiments}
\begin{figure*}[h!]
  \centering
	\begin{subfigure}{0.24\textwidth}
    \centering
    \includegraphics[width=\textwidth]{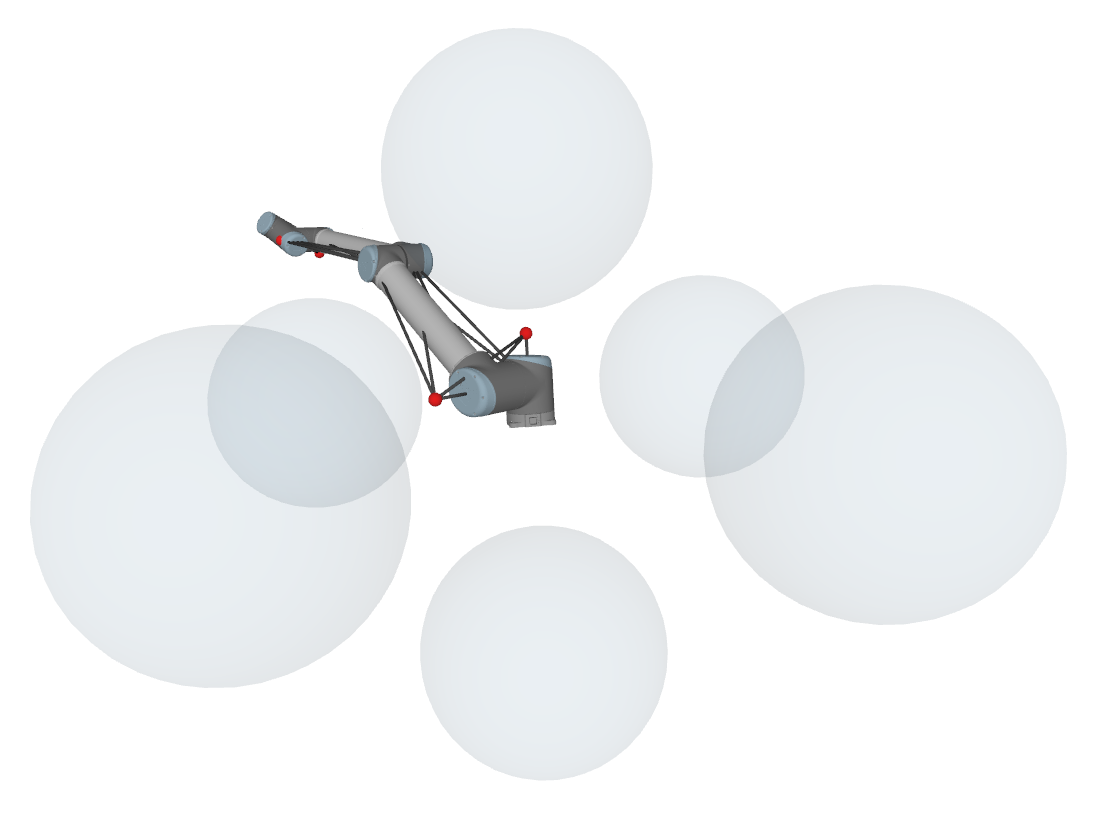}
    \caption{\texttt{UR10}, (\textit{octahedron}).}\label{fig:urdf-ur10}
  \end{subfigure}
	\begin{subfigure}{0.24\textwidth}
    \centering
    \includegraphics[width=0.7\textwidth]{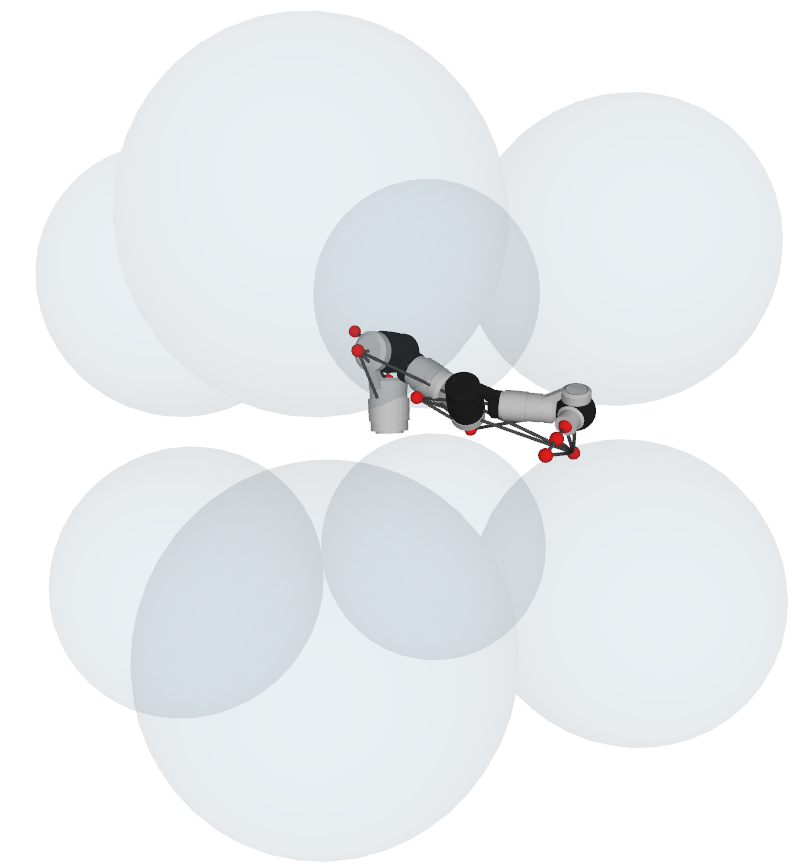}
    \caption{\texttt{Schunk LWA4D}, (\textit{cube}).}\label{fig:urdf-lwa4d}
  \end{subfigure}
	\begin{subfigure}{0.24\textwidth}
    \centering
    \includegraphics[width=0.8\textwidth]{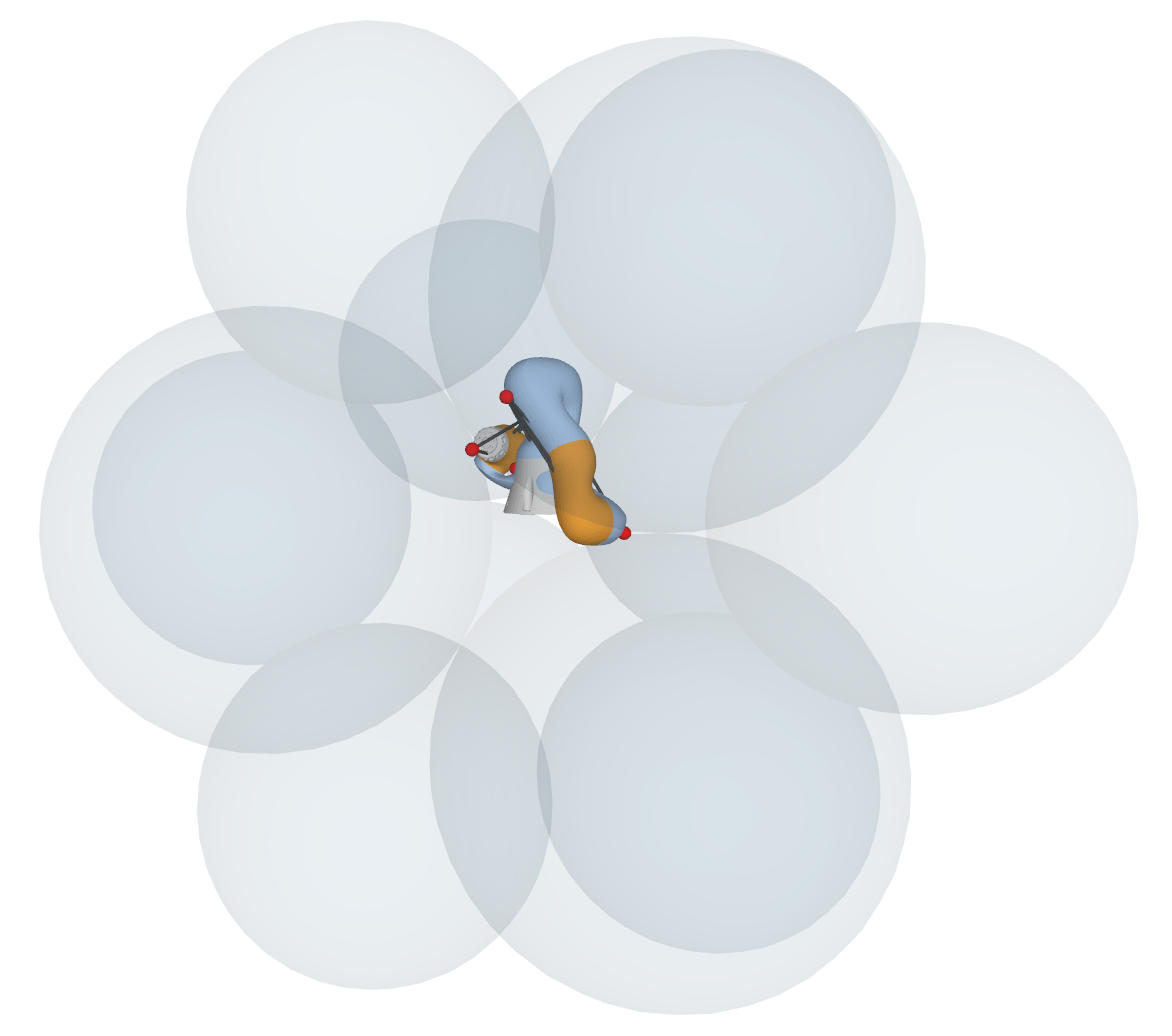}
    \caption{\texttt{KUKA IIWA}, (\textit{icosahedron}).}\label{fig:urdf-kuka}
  \end{subfigure}
  \begin{subfigure}{0.24\textwidth}
  	\centering
    \includegraphics[width=0.8\textwidth]{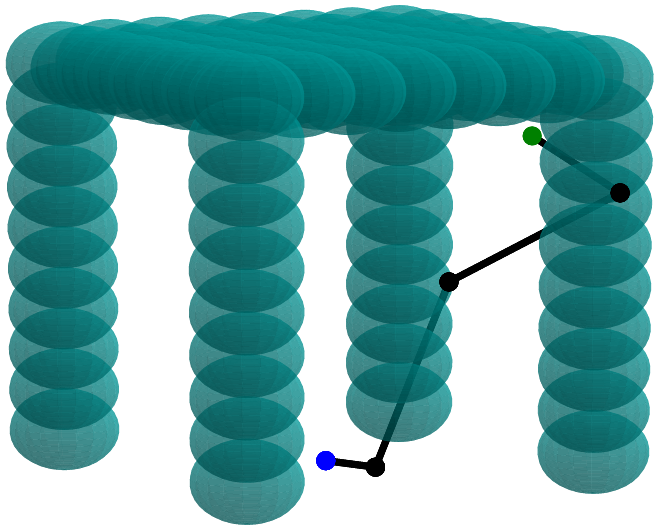}
    \caption{\texttt{9-DOF}, (\textit{table}).}\label{fig:table-9_dof}
  \end{subfigure}
\caption{Four robot manipulators (and obstacle configurations) used in our experiments. For (a)-(c), we also visualize the edges (dark grey lines) and nodes (red) of the associated acyclic graph used to form our distance-geometric model (cf. \Cref{fig:revolute_obstacles}). These three environments, labeled \textit{octahedron}, \textit{cube} and \textit{icosahedron}, correspond to obstacle configurations defined by the vertices of the respective Platonic solid centred at the base of the robot.}
\label{fig:urdf_fig}
\vspace{-1mm}
\end{figure*}

We evaluate our proposed approach on IK problems for the three commercial robots shown in \Cref{fig:urdf_fig} in a variety of environments.
In all cases, we generate \emph{feasible} IK problems by taking uniform random sample configurations $\Matrix{\Theta} \in \Configuration$, rejecting configurations that violate collision avoidance constraints,\footnote{Note that for redundant manipulators this procedure can reject some feasible goals, since (infinitely) many other configurations may exist that reach the desired end-effector pose without any collisions.} and using the resulting end-effector pose $F(\Matrix{\Theta})$ as a goal.
Each problem instance is solved with \Cref{alg:convex_iteration} (\Ours), with the MOSEK interior point solver~\cite{andersen_mosek_2000} used for each iteration of \Cref{prob:rank_min_sdp}.
To determine whether \Cref{alg:convex_iteration}'s solution to a particular IK problem is successful, we first use the procedure in \cite{maric_riemannian_2021} to reconstruct the joint configuration $\Matrix{\Theta}$ from points $\Matrix{X}$ extracted from the rank-$\Dim$ matrix $\Matrix{Z}^\star$ returned after a maximum of 10 iterations of the algorithm in \Cref{sec:convex_iteration}. 
This joint configuration is treated as \Ours's solution and fed into the forward kinematics in \Cref{eq:forward_kinematics} to obtain the end-effector pose and any workspace constraint violations.
A solution is considered correct when it satisfies obstacle constraints to within a \mbox{$0.01$ m} tolerance, has an end-effector position error lower than \mbox{$0.01$ m}, and has an orientation error lower than $0.01$ rad ($0.6^\circ$).

In order to evaluate the advantages of our approach over formulations based on joint angles, we also implement an IK solver that uses nonlinear optimization.
Namely, we use the square of the end-effector pose error, $\Vector{e}$, as the objective function of a nonlinear program, with collision avoidance constraints equivalent to those used by \Ours represented as nonlinear inequality constraints:
\begin{align} \label{prob:slsqp}
\min_{\Vector{\Theta} \in \Configuration} &\quad \Norm{\Vector{e}\left(\Vector{\Theta}, \Matrix{T}^{\mathrm{g}}\right)}^2 \\
\text{\emph{s.t.}} &\quad \Norm{\Vector{x}_i(\Vector{\Theta}) - \Vector{c}_j}^2 \geq l^2_j \enspace \forall i \in \JointVertices, \enspace \forall j \in \Obstacles,\notag
\end{align}
where the error $\Vector{e}$ between the current and goal poses is represented as a twist using the logarithmic map~\cite{lynch2017modern}.
The spherical obstacles in $\Obstacles$ are parameterized as in \Cref{eq:spherical_obstacle}.
This formulation can be solved using Sequential Quadratic Programming (SQP) and has previously been used within the TRAC-IK algorithm \cite{beeson2015trac}.
Since TRAC-IK does not support obstacles, we implement our solver using the \SLSQP routine in the \texttt{scipy} Python package~\cite{virtanen2020scipy}.
As an additional comparison, we solve~\Cref{prob:slsqp} using \IPOPT, a highly efficient implementation of an interior-point filter line-search algorithm~\cite{wachter2006implementation}.
All experiments are performed on a laptop with an Intel i7-8750H CPU running at 2.20 GHz and with 16 GB of RAM.


\begin{table*}
\footnotesize
\centering
\resizebox{0.98\textwidth}{!}{
\begin{tabular}{lcccccccccccc}
    Env. & \multicolumn{3}{c}{Free} & \multicolumn{3}{c}{Octahedron} & \multicolumn{3}{c}{Cube} & \multicolumn{3}{c}{Icosahedron}\\
    \cmidrule(r){2-4}  \cmidrule(r){5-7}   \cmidrule(r){8-10} \cmidrule(r){11-13}
     &\texttt{SLSQP}&\texttt{IPOPT} &\texttt{CIDGIK}&\texttt{SLSQP}&\texttt{IPOPT} &\texttt{CIDGIK}&\texttt{SLSQP}&\texttt{IPOPT} &\texttt{CIDGIK}&\texttt{SLSQP}&\texttt{IPOPT} &\texttt{CIDGIK}\\
    \midrule
    \midrule
    UR10  & 89.9 $\pm$ 1.1 & \textbf{100 $\pm$ 0.0} & 90.8 $\pm$ 1.0 & 56.3 $\pm$ 1.8 & 47.8 $\pm$ 1.8 & \textbf{76.2 $\pm$ 1.5} & 57.9 $\pm$ 1.8 & 46.7 $\pm$ 1.8 & \textbf{67.1 $\pm$ 1.7} & 15.2 $\pm$ 1.3 & 15.4 $\pm$ 1.3 & \textbf{51.6 $\pm$ 1.8} \\
    KUKA & 98.1 $\pm$ 0.5 & \textbf{99.9 $\pm$ 0.1} & 99.7 $\pm$ 0.2 & 97.6 $\pm$ 0.5 & 66.6 $\pm$ 1.7 & \textbf{99.8 $\pm$ 0.2} & 95.5 $\pm$ 0.7 & 61.4 $\pm$ 1.7 & \textbf{99.2 $\pm$ 0.3} & 93.1 $\pm$ 0.9 & 64.4 $\pm$ 1.7 & \textbf{99.6 $\pm$ 0.2}\\
    LWA4D & 97.0 $\pm$ 0.6 & \textbf{100 $\pm$ 0.0} & 99.5 $\pm$ 0.3 & 95.6 $\pm$ 0.7 & 52.2 $\pm$ 1.8 & \textbf{99.8 $\pm$ 0.2} & 95.6 $\pm$ 0.7 & 53.4 $\pm$ 1.8 & \textbf{99.7 $\pm$ 0.2} & 93.6 $\pm$ 0.9 & 53.3 $\pm$ 1.8 & \textbf{99.7 $\pm$ 0.2}\\
\end{tabular}
}
\caption{Percentages of successfully solved problems as 95\% Jeffreys confidence intervals~\cite{tonycai_one-sided_2005}.}\label{tab:success_comp}
\end{table*}

\begin{table*}
\vspace{-1.5mm}
\footnotesize
\centering
\resizebox{0.98\textwidth}{!}{
\begin{tabular}{lcccccccccccc}
    Env. & \multicolumn{3}{c}{Free} & \multicolumn{3}{c}{Octahedron} & \multicolumn{3}{c}{Cube} & \multicolumn{3}{c}{Icosahedron}\\
    \cmidrule(r){2-4}  \cmidrule(r){5-7}   \cmidrule(r){8-10} \cmidrule(r){11-13}
     &\texttt{SLSQP}&\texttt{IPOPT} &\texttt{CIDGIK}&\texttt{SLSQP}&\texttt{IPOPT} &\texttt{CIDGIK}&\texttt{SLSQP}&\texttt{IPOPT} &\texttt{CIDGIK}&\texttt{SLSQP}&\texttt{IPOPT} &\texttt{CIDGIK}\\

    \midrule
    \midrule
    UR10  & 0.07 (0.02) & \textbf{0.01 (0.00)} & 0.24 (0.20) & 0.36 (0.14) & 2.3 (2.1) & \textbf{0.16 (0.11)} & 0.43 (0.17) & 2.6 (2.1) & \textbf{0.18 (0.12)} & 0.40 (0.19) & 1.6 (1.8) & \textbf{0.20 (0.12)} \\
    KUKA & \textbf{0.09 (0.04)} & 0.74 (0.72) & 0.16 (0.07) & 0.58 (0.24) & 4.0 (2.4) & \textbf{0.14 (0.07)} & 0.64 (0.26) & 4.7 (2.5) & \textbf{0.14 (0.08)} & 0.87 (0.40) & 4.6 (2.5) & \textbf{0.14 (0.07)} \\
    LWA4D & \textbf{0.10 (0.04)} & 0.88 (0.78) & 0.15 (0.07) & 0.58 (0.22) & 5.1 (2.1) & \textbf{0.13 (0.06)} & 0.64 (0.25) & 5.2 (2.1) & \textbf{0.13 (0.06)} & 0.91 (0.46) & 12.8 (5.4) & \textbf{0.13 (0.06)} \\
\end{tabular}
}
\caption{Mean, with standard deviation in brackets, of solution times in seconds. For each algorithm, ``solution time" does not include problem setup time. For \Ours, the sum of solver times returned by each iteration of both \Cref{prob:rank_min_sdp} and \Cref{prob:fantope_sdp} is used.}\label{tab:time_comp}
\vspace{-1mm}
\end{table*}

\begin{figure*}
\centering 
\includegraphics[width=\textwidth]{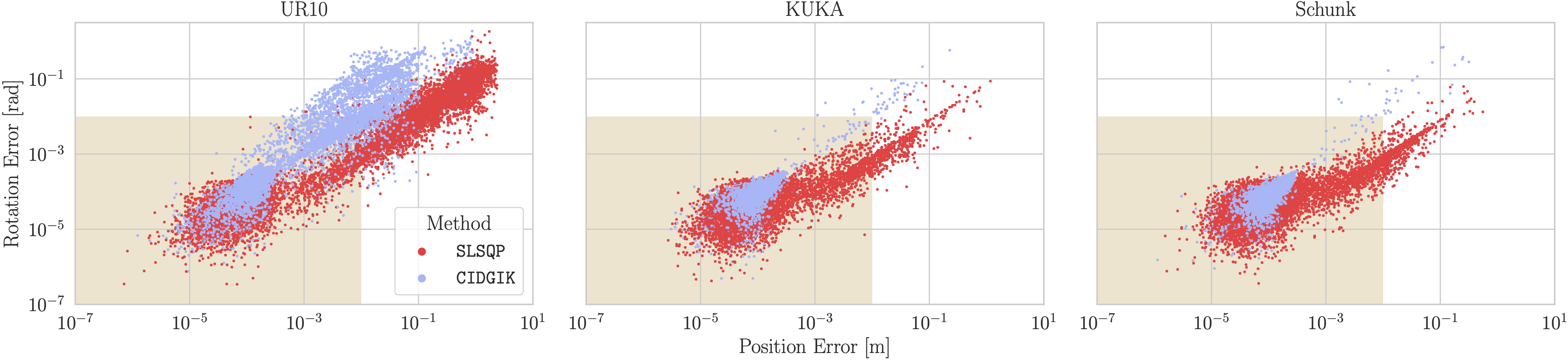}
\caption{A comparison of the end-effector position and orientation errors. Each sub-plot contains problem instances from all four obstacle environments. The shaded rectangle indicates the region of error tolerance (i.e., contains all successful runs).}
\label{fig:error_scatter}
\vspace{-5mm}
\end{figure*}

\begin{figure*}
\centering 
\includegraphics[width=\textwidth]{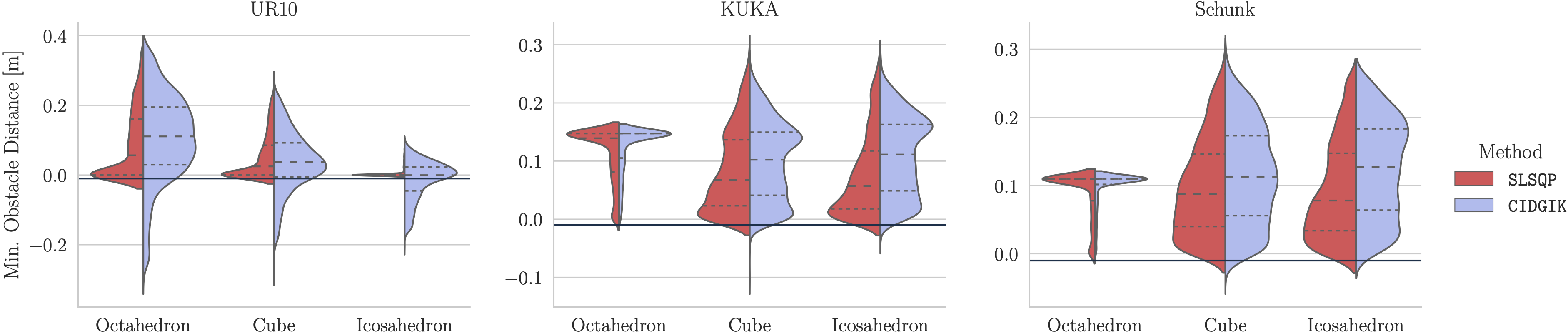}
\caption{Violin plots describing the distribution of the minimum distance to an obstacle for different algorithms, robots, and environments. The dashed lines are the quartile boundaries of each distribution, and the solid dark line is the error tolerance (0.01 m). Note that the performance of \SLSQP on the UR10 in the \textit{icosahedron} environment has extremely low variance around zero.}
\label{fig:obstacle_violins}
\end{figure*}

To provide a quantitative evaluation of our approach, we generate and solve 3,000 problem instances for each of the three robots depicted in Figures \ref{fig:urdf-ur10}-\ref{fig:urdf-lwa4d} in each of the distinct environments depicted in those same figures, as well as a fourth obstacle-free environment.
For consistency, all three algorithms only consider collisions between joint locations and obstacles (i.e., auxiliary points are not added along the links in \Ours).
Tables \ref{tab:success_comp} and \ref{tab:time_comp} summarize respectively the success rate and runtime of each algorithm for each robot-environment pair.

Comparing success rates in \Cref{tab:success_comp} reveals that our method solves a larger percentage of problems overall than both \SLSQP and \IPOPT in all environments featuring obstacles, achieving a success rate of $>99$\% in many instances.
Likewise, \Ours's runtime in \Cref{tab:time_comp} is consistently lower than its competitors' across all three obstacle-filled environments. 
In the obstacle-free case, \SLSQP runs faster than \Ours with similar accuracy, and \IPOPT achieves a perfect success rate and extremely low mean runtime on the UR10, which the other two algorithms struggle with.
Curiously, while its competitors slow down significantly with the addition of obstacles, the solution times for \Ours barely increase (or even slightly \textit{decrease}) as the number of obstacles grows.
In future work, we plan to investigate whether this trend holds for other more complex robots, and for environments with many hundreds or thousands of obstacles.

\Cref{fig:error_scatter} displays the distribution of position and rotation errors for \Ours and \SLSQP on each robot, aggregated across experiments from all four environments.
\Cref{fig:obstacle_violins} depicts the distribution of the distance to the nearest obstacle across experiments.
Together, these two figures reveal that with the exception of \Ours on the UR10 robot, most failures were in fact due to end-effector error as opposed to obstacle constraint violations.
Finally, \Cref{tab:9_dof_results} summarizes 1,000 experiments comparing \Ours and \SLSQP on a hyper-redundant 9-DOF manipulator~\cite{xu_design_2014} in the 100-obstacle \textit{table} environment displayed in \Cref{fig:table-9_dof}.
These results demonstrate \Ours's statistically significant superiority over \SLSQP, both in terms of accuracy and runtime, as both the number of obstacles and robot DOF increase.

\begin{table}[h!]
\footnotesize
\centering
\begin{tabular}{l | cccc}
\scriptsize{Method} & \scriptsize{Pos. Err. [cm]} & \scriptsize{Rot. Err. [mrad]} & \scriptsize{Sol. Time [s]} & \scriptsize{Succ. [\%]} \\
\midrule
\SLSQP & 1.6 (12) & 1 (14) & 1.5 (0.7) & 97 $\pm$ 1.1 \\
\Ours & 0.02 (0.21) & 0.13 (2) & 0.16 (0.1) & 99 $\pm$ 0.4
\end{tabular}
\caption{Results for 1,000 9-DOF manipulator experiments in the \textit{table} environment (\Cref{fig:table-9_dof}). Errors and and solve times are reported as mean with standard deviation in brackets, while the success rates are given as 95\% Jeffreys confidence intervals~\cite{tonycai_one-sided_2005}.}
\label{tab:9_dof_results}
\vspace{-6mm}
\end{table}

\section{Conclusions and Future Work} \label{sec:conclusion}
We have presented a novel distance-geometric approach to solving inverse kinematics problems involving redundant manipulators with arbitrary spherical and planar workspace constraints.
Preliminary experiments demonstrate that our algorithm significantly outperforms benchmark algorithms in obstacle-laden environments.
Crucially, our problem formulation connects IK to the rich literature on SDP relaxations for distance geometry problems, providing us with the novel and elegant geometric interpretation of IK discussed in \Cref{sec:geometric_interpretation}.

Future work will involve implementing an SDP solver that exploits the chordal sparsity of robot kinematics, and testing \Ours in more complex scenarios. 
While our model is quite general, it still needs to be extended to incorporate arbitrary joint angle limits.
Moreover, robots with successive joint axes that are not coplanar need additional variables and constraints to ensure that non-physical reflections of parts of a robot are not included in the feasible set of \Cref{prob:feasibility_sdp}. 
These modifications can be incorporated into our framework~\cite{giamou_supplementary_2022}, but their effect on accuracy and runtime remains to be seen.
Finally, while our method uses global optimization to solve the \textit{subproblem} in each iteration, it remains to be shown whether global convergence guarantees exist for \Ours.
These guarantees may depend on robot structure or hyperparameter settings (e.g., a particular choice of $\Matrix{C}^{\{0\}} \neq \Identity$), and modifications might affect which of the possibly infinite feasible solutions \Ours returns.
Understanding this behaviour is key to making \Ours a fast and reliable subroutine for a variety of challenging motion planning applications.

\balance
\bibliographystyle{IEEEtran}
\bibliography{robotics_abbrv, references}

\begin{thebibliography}{10}
\providecommand{\url}[1]{#1}
\csname url@rmstyle\endcsname
\providecommand{\newblock}{\relax}
\providecommand{\bibinfo}[2]{#2}
\providecommand\BIBentrySTDinterwordspacing{\spaceskip=0pt\relax}
\providecommand\BIBentryALTinterwordstretchfactor{4}
\providecommand\BIBentryALTinterwordspacing{\spaceskip=\fontdimen2\font plus
\BIBentryALTinterwordstretchfactor\fontdimen3\font minus
  \fontdimen4\font\relax}
\providecommand\BIBforeignlanguage[2]{{%
\expandafter\ifx\csname l@#1\endcsname\relax
\typeout{** WARNING: IEEEtran.bst: No hyphenation pattern has been}%
\typeout{** loaded for the language `#1'. Using the pattern for}%
\typeout{** the default language instead.}%
\else
\language=\csname l@#1\endcsname
\fi
#2}}

\bibitem{porta_inverse_2005}
J.~M. Porta, L.~Ros, and F.~Thomas, ``{Inverse Kinematics by Distance Matrix
  Completion},'' in \emph{Proc. of 12th Int. Workshop on Computational
  Kinematics}.\hskip 1em plus 0.5em minus 0.4em\relax Elsevier, 2005.

\bibitem{Le_Naour_2019}
T.~Le~Naour, N.~Courty, and S.~Gibet, ``{Kinematics in the Metric Space},''
  \emph{Computers \& Graphics}, vol.~84, p. 13–23, Nov. 2019.

\bibitem{blanchini_convex_2017}
F.~Blanchini, G.~Fenu, G.~Giordano, and F.~A. Pellegrino,
  ``\BIBforeignlanguage{en}{{A Convex Programming Approach to the Inverse
  Kinematics Problem for Manipulators under Constraints}},''
  \emph{\BIBforeignlanguage{en}{European J. of Control}}, vol.~33, pp. 11--23,
  Jan. 2017.

\bibitem{dattorro_convex_2005}
J.~Dattorro, \emph{\BIBforeignlanguage{en}{Convex Optimization \& Euclidean
  Distance Geometry}}.\hskip 1em plus 0.5em minus 0.4em\relax {Palo Alto,
  California}: MeBoo Publishing, 2005.

\bibitem{spong2020robot}
M.~W. Spong, S.~Hutchinson, and M.~Vidyasagar, \emph{{Robot Modeling and
  Control}}.\hskip 1em plus 0.5em minus 0.4em\relax John Wiley \& Sons, 2005.

\bibitem{manocha_efficient_1994}
D.~Manocha and J.~Canny, ``\BIBforeignlanguage{en}{{Efficient Inverse
  Kinematics for General {{6R}} Manipulators}},''
  \emph{\BIBforeignlanguage{en}{IEEE Trans. on Robotics and Automation}},
  vol.~10, no.~5, pp. 648--657, 1994.

\bibitem{buss_introduction_2004}
\BIBentryALTinterwordspacing
S.~R. Buss, ``Introduction to {{Inverse Kinematics}} with {{Jacobian
  Transpose}}, {{Pseudoinverse}} and {{Damped Least Squares}} methods,'' 2009.
  [Online]. Available: \url{https://mathweb.ucsd.edu/~sbuss/ResearchWeb/}
\BIBentrySTDinterwordspacing

\bibitem{aristidouInverseKinematicsTechniques2018}
A.~Aristidou, J.~Lasenby, Y.~Chrysanthou, and A.~Shamir,
  ``\BIBforeignlanguage{en}{Inverse {{Kinematics Techniques}} in {{Computer
  Graphics}}: {{A Survey}}},'' \emph{\BIBforeignlanguage{en}{Computer Graphics
  Forum}}, vol.~37, no.~6, pp. 35--58, Sept. 2018.

\bibitem{beeson2015trac}
P.~Beeson and B.~Ames, ``{TRAC-IK}: {An Open-Source Library for Improved
  Solving of Generic Inverse Kinematics},'' in \emph{15th Int. Conf. on
  Humanoid Robots (Humanoids)}, 2015, pp. 928--935.

\bibitem{dai_global_2019}
H.~Dai, G.~Izatt, and R.~Tedrake, ``\BIBforeignlanguage{en}{{Global Inverse
  Kinematics via Mixed-Integer Convex Optimization}},''
  \emph{\BIBforeignlanguage{en}{Intl. J. Robotics Research}}, vol.~38, no.
  12-13, pp. 1420--1441, Oct. 2019.

\bibitem{libertiEuclideanDistanceGeometry2014}
L.~Liberti, C.~Lavor, N.~Maculan, and A.~Mucherino,
  ``\BIBforeignlanguage{en}{Euclidean {{Distance Geometry}} and
  {{Applications}}},'' \emph{\BIBforeignlanguage{en}{SIAM Rev.}}, vol.~56,
  no.~1, pp. 3--69, Jan. 2014.

\bibitem{dokmanic_euclidean_2015}
I.~Dokmanic, R.~Parhizkar, J.~Ranieri, and M.~Vetterli,
  ``\BIBforeignlanguage{en}{Euclidean {{Distance Matrices}}: {{Essential
  Theory}}, {{Algorithms}} and {{Applications}}},''
  \emph{\BIBforeignlanguage{en}{IEEE Signal Processing Magazine}}, vol.~32,
  no.~6, pp. 12--30, Nov. 2015.

\bibitem{porta_branch-and-prune_2005}
J.~Porta, L.~Ros, F.~Thomas, and C.~Torras, ``\BIBforeignlanguage{en}{{A
  Branch-and-Prune Solver for Distance Constraints}},''
  \emph{\BIBforeignlanguage{en}{{IEEE} Trans. Robotics}}, vol.~21, no.~2, pp.
  176--187, Apr. 2005.

\bibitem{blanchini_inverse_2015}
F.~Blanchini, G.~Fenu, G.~Giordano, and F.~A. Pellegrino,
  ``\BIBforeignlanguage{en}{{Inverse Kinematics by Means of Convex
  Programming}: {{Some}} {Developments}},'' in
  \emph{\BIBforeignlanguage{en}{{{IEEE Intl. Conf.}} on {{Automation Science}}
  and {{Engineering}}}}, {Gothenburg, Sweden}, Aug. 2015, pp. 515--520.

\bibitem{so_theory_2007}
A.~M.-C. So and Y.~Ye, ``\BIBforeignlanguage{en}{{Theory of Semidefinite
  Programming} for {{Sensor Network Localization}}},''
  \emph{\BIBforeignlanguage{en}{Mathematical Programming}}, vol. 109, no. 2-3,
  pp. 367--384, Jan. 2007.

\bibitem{cifuentes_local_2020}
D.~Cifuentes, S.~Agarwal, P.~A. Parrilo, and R.~R. Thomas,
  ``\BIBforeignlanguage{en}{{On the Local Stability of Semidefinite
  Relaxations}},'' \emph{\BIBforeignlanguage{en}{arXiv:1710.04287 [math]}},
  Aug. 2020.

\bibitem{boyd_semidefinite_1997}
S.~Boyd and L.~Vandenberghe, ``\BIBforeignlanguage{en}{Semidefinite
  {{Programming Relaxations}} of {{Non}}-{{Convex Problems}} in {{Control}} and
  {{Combinatorial Optimization}}},'' in
  \emph{\BIBforeignlanguage{en}{Communications, {{Computation}}, {{Control}},
  and {{Signal Processing}}}}, 1997, pp. 279--287.

\bibitem{majumdar_survey_2019}
A.~Majumdar, G.~Hall, and A.~A. Ahmadi, ``\BIBforeignlanguage{en}{A {{Survey}}
  of {{Recent Scalability Improvements}} for {{Semidefinite Programming}} with
  {{Applications}} in {{Machine Learning}}, {{Control}}, and {{Robotics}}},''
  \emph{\BIBforeignlanguage{en}{arXiv:1908.05209 [cs, eess, math]}}, Sept.
  2019.

\bibitem{trutman_globally_2020}
P.~Trutman, S.~E.~D. Mohab, D.~Henrion, and T.~Pajdla,
  ``\BIBforeignlanguage{en}{Globally {{Optimal Solution}} to {{Inverse
  Kinematics}} of {{7DOF Serial Manipulator}}},''
  \emph{\BIBforeignlanguage{en}{arXiv:2007.12550 [cs, math]}}, July 2020.

\bibitem{maric_riemannian_2021}
F.~Mari{\'c}, M.~Giamou, A.~W. Hall, S.~Khoubyarian, I.~Petrovi{\'c}, and
  J.~Kelly, ``\BIBforeignlanguage{en}{Riemannian {{Optimization}} for
  {{Distance-Geometric Inverse Kinematics}}},''
  \emph{\BIBforeignlanguage{en}{{IEEE} Trans. Robotics}}, pp. 1--20, Dec. 2021.

\bibitem{maricInverseKinematicsSerial2020}
F.~Mari{\'c}, M.~Giamou, S.~Khoubyarian, I.~Petrovi{\'c}, and J.~Kelly,
  ``{Inverse Kinematics for Serial Kinematic Chains via Sum of Squares
  Optimization},'' in \emph{IEEE Intl. Conf. Robotics and Automation (ICRA)},
  Aug. 2020, pp. 7101--7107.

\bibitem{nie_sum_2009}
J.~Nie, ``\BIBforeignlanguage{en}{{Sum of Squares Method for Sensor Network
  Localization}},'' \emph{\BIBforeignlanguage{en}{Computational Optimization
  and Applications}}, vol.~43, no.~2, pp. 151--179, June 2009.

\bibitem{dingSensorNetworkLocalization2010}
Y.~Ding, N.~Krislock, J.~Qian, and H.~Wolkowicz,
  ``\BIBforeignlanguage{en}{Sensor {{Network Localization}}, {{Euclidean
  Distance Matrix}} {Completions, and Graph Realization}},''
  \emph{\BIBforeignlanguage{en}{Optim. Eng.}}, vol.~11, no.~1, pp. 45--66, Feb.
  2010.

\bibitem{yenamandra_convex_2019}
T.~Yenamandra, F.~Bernard, J.~Wang, F.~Mueller, and C.~Theobalt,
  ``\BIBforeignlanguage{en}{Convex {{Optimisation}} for {{Inverse
  Kinematics}}},'' \emph{\BIBforeignlanguage{en}{2019 Intl. Conf. on 3D Vision
  (3DV)}}, pp. 318--327, Sept. 2019.

\bibitem{erleben_solving_2019}
K.~Erleben and S.~Andrews, ``\BIBforeignlanguage{en}{{Solving Inverse
  Kinematics Using Exact Hessian Matrices}},''
  \emph{\BIBforeignlanguage{en}{Computers \& Graphics}}, vol.~78, pp. 1--11,
  Feb. 2019.

\bibitem{giamou_supplementary_2022}
M.~Giamou, F.~Mari{\'c}, D.~M. Rosen, V.~Peretroukhin, N.~Roy, I.~Petrovi{\'c},
  and J.~Kelly, ``{Supplementary Material for Convex Iteration for
  Distance-Geometric Inverse Kinematics},'' \emph{arXiv:2109.03374 [cs]}, 2022.

\bibitem{varava_caging_2020}
A.~Varava, J.~F. Carvalho, F.~T. Pokorny, and D.~Kragic,
  ``\BIBforeignlanguage{en}{Caging and {{Path Non}}-{Existence}: {{A
  Deterministic Sampling}}-{{Based Verification Algorithm}}},'' in
  \emph{\BIBforeignlanguage{en}{Robotics {{Research}}}}, 2020, vol.~10, pp.
  589--604.

\bibitem{akin_computing_2015}
R.~Deits and R.~Tedrake, ``\BIBforeignlanguage{en}{Computing {{Large Convex
  Regions}} of {{Obstacle}}-{{Free Space Through Semidefinite Programming}}},''
  in \emph{\BIBforeignlanguage{en}{Algorithmic {{Foundations}} of {{Robotics
  XI}}}}, Apr. 2015, vol. 107, pp. 109--124.

\bibitem{krislock_explicit_2010}
N.~Krislock and H.~Wolkowicz, ``\BIBforeignlanguage{en}{Explicit {{Sensor
  Network Localization}} using {{Semidefinite Representations}} and {{Facial
  Reductions}}},'' \emph{\BIBforeignlanguage{en}{SIAM J. on Optimization}},
  vol.~20, no.~5, pp. 2679--2708, Jan. 2010.

\bibitem{aspnes_computational_2004}
J.~Aspnes, D.~Goldenberg, and Y.~R. Yang, ``\BIBforeignlanguage{en}{On the
  {{Computational Complexity}} of {{Sensor Network Localization}}},'' in
  \emph{\BIBforeignlanguage{en}{Algorithmic {{Aspects}} of {{Wireless Sensor
  Networks}}}}, 2004, pp. 32--44.

\bibitem{boyd2004convex}
S.~Boyd and L.~Vandenberghe, \emph{Convex Optimization}.\hskip 1em plus 0.5em
  minus 0.4em\relax Cambridge University Press, 2004.

\bibitem{recht_guaranteed_2010}
B.~Recht, M.~Fazel, and P.~A. Parrilo, ``\BIBforeignlanguage{en}{Guaranteed
  {{Minimum}}-{{Rank Solutions}} of {{Linear Matrix Equations}} via {{Nuclear
  Norm Minimization}}},'' \emph{\BIBforeignlanguage{en}{SIAM Review}}, vol.~52,
  no.~3, pp. 471--501, Jan. 2010.

\bibitem{chandrasekaran_convex_2012}
V.~Chandrasekaran, B.~Recht, P.~A. Parrilo, and A.~S. Willsky,
  ``\BIBforeignlanguage{en}{The {{Convex Geometry}} of {{Linear Inverse
  Problems}}},'' \emph{\BIBforeignlanguage{en}{Found. Comput. Math}}, pp.
  805--849, Oct. 2012.

\bibitem{wang_chordal_2018}
W.~Wang and N.~Yu, ``\BIBforeignlanguage{en}{Chordal {{Conversion Based Convex
  Iteration Algorithm}} for {{Three}}-{{Phase Optimal Power Flow Problems}}},''
  \emph{\BIBforeignlanguage{en}{IEEE Trans. on Power Systems}}, vol.~33, no.~2,
  pp. 1603--1613, Mar. 2018.

\bibitem{andersen_mosek_2000}
E.~D. Andersen and K.~D. Andersen, ``\BIBforeignlanguage{en}{The {{MOSEK
  Interior Point Optimizer}} for {{Linear Programming}}: {{An Implementation}}
  of the {{Homogeneous Algorithm}}},'' in \emph{\BIBforeignlanguage{en}{High
  {{Performance Optimization}}}}, {Boston, MA}, 2000, vol.~33, pp. 197--232.

\bibitem{lynch2017modern}
K.~M. Lynch and F.~C. Park, \emph{Modern Robotics}.\hskip 1em plus 0.5em minus
  0.4em\relax Cambridge University Press, 2017.

\bibitem{virtanen2020scipy}
P.~Virtanen \emph{et~al.}, ``{SciPy 1.0: Fundamental Algorithms for Scientific
  Computing} in {Python},'' \emph{Nature Methods}, vol.~17, no.~3, pp.
  261--272, 2020.

\bibitem{wachter2006implementation}
A.~W{\"a}chter and L.~T. Biegler, ``{On the Implementation of an Interior-Point
  Filter Line-Search Algorithm for Large-Scale Nonlinear Programming},''
  \emph{Mathematical programming}, vol. 106, no.~1, pp. 25--57, 2006.

\bibitem{tonycai_one-sided_2005}
T.~Tony~Cai, ``\BIBforeignlanguage{en}{{One-Sided Confidence Intervals in
  Discrete Distributions}},'' \emph{\BIBforeignlanguage{en}{J. of Statistical
  Planning and Inference}}, vol. 131, no.~1, pp. 63--88, Apr. 2005.

\bibitem{xu_design_2014}
X.~Xu, H.~Ananthanarayanan, and R.~Ordonez, ``{Design and Construction} of
  9-{{DOF}} {Hyper-Redundant Robotic Arm},'' in \emph{{{IEEE National
  Aerospace}} and {{Electronics Conference}}}, June 2014, pp. 321--326.

\end{thebibliography}


\begin{thebibliography}{1}
\providecommand{\url}[1]{#1}
\csname url@rmstyle\endcsname
\providecommand{\newblock}{\relax}
\providecommand{\bibinfo}[2]{#2}
\providecommand\BIBentrySTDinterwordspacing{\spaceskip=0pt\relax}
\providecommand\BIBentryALTinterwordstretchfactor{4}
\providecommand\BIBentryALTinterwordspacing{\spaceskip=\fontdimen2\font plus
\BIBentryALTinterwordstretchfactor\fontdimen3\font minus
  \fontdimen4\font\relax}
\providecommand\BIBforeignlanguage[2]{{%
\expandafter\ifx\csname l@#1\endcsname\relax
\typeout{** WARNING: IEEEtran.bst: No hyphenation pattern has been}%
\typeout{** loaded for the language `#1'. Using the pattern for}%
\typeout{** the default language instead.}%
\else
\language=\csname l@#1\endcsname
\fi
#2}}

\bibitem{odonoghue_scs_2016}
B.~O'Donoghue, E.~Chu, N.~Parikh, and S.~Boyd, ``Conic optimization via
  operator splitting and homogeneous self-dual embedding,'' \emph{Journal of
  Optimization Theory and Applications}, vol. 169, no.~3, pp. 1042--1068, June
  2016.

\bibitem{andersen_mosek_2000}
E.~D. Andersen and K.~D. Andersen, ``\BIBforeignlanguage{en}{The {{MOSEK
  Interior Point Optimizer}} for {{Linear Programming}}: {{An Implementation}}
  of the {{Homogeneous Algorithm}}},'' in \emph{\BIBforeignlanguage{en}{High
  {{Performance Optimization}}}}, {Boston, MA}, 2000, vol.~33, pp. 197--232.

\bibitem{diamond_cvxpy_2016}
S.~Diamond and S.~Boyd, ``Cvxpy: A python-embedded modeling language for convex
  optimization,'' \emph{The Journal of Machine Learning Research}, vol.~17,
  no.~1, pp. 2909--2913, 2016.

\bibitem{majumdar_survey_2019}
A.~Majumdar, G.~Hall, and A.~A. Ahmadi, ``\BIBforeignlanguage{en}{A {{Survey}}
  of {{Recent Scalability Improvements}} for {{Semidefinite Programming}} with
  {{Applications}} in {{Machine Learning}}, {{Control}}, and {{Robotics}}},''
  \emph{\BIBforeignlanguage{en}{arXiv:1908.05209 [cs, eess, math]}}, Sept.
  2019.

\bibitem{wang_chordal_2018}
W.~Wang and N.~Yu, ``\BIBforeignlanguage{en}{Chordal {{Conversion Based Convex
  Iteration Algorithm}} for {{Three}}-{{Phase Optimal Power Flow Problems}}},''
  \emph{\BIBforeignlanguage{en}{IEEE Trans. on Power Systems}}, vol.~33, no.~2,
  pp. 1603--1613, Mar. 2018.

\bibitem{garstka_cosmo_2019}
M.~Garstka, M.~Cannon, and P.~Goulart, ``\BIBforeignlanguage{en}{{{COSMO}}:
  {{A}} conic operator splitting method for large convex problems},'' in
  \emph{\BIBforeignlanguage{en}{2019 18th {{European Control Conference}}
  ({{ECC}})}}, {Naples, Italy}, June 2019, pp. 1951--1956.

\bibitem{maricInverseKinematicsSerial2020}
F.~Mari{\'c}, M.~Giamou, S.~Khoubyarian, I.~Petrovi{\'c}, and J.~Kelly,
  ``{Inverse Kinematics for Serial Kinematic Chains via Sum of Squares
  Optimization},'' in \emph{IEEE Intl. Conf. Robotics and Automation (ICRA)},
  Aug. 2020, pp. 7101--7107.

\end{thebibliography}
\end{document}


\title{Supplementary Material for ``Convex Iteration for Distance-Geometric Inverse Kinematics"}

\author{Matthew Giamou, Filip Mari\'c, David M. Rosen, Valentin Peretroukhin,\\ Nicholas Roy, Ivan Petrovi\'c, and Jonathan Kelly

}

\maketitle

\begin{abstract}
Supplementary material for the 2021 IEEE Robotics and Automation Letters submission entitled ``Convex Iteration for Distance Geometric Inverse Kinematics".
\end{abstract}


\section{Mathematics}
In this section, we describe some of the mathematics behind our formulation of the SDP \Ours solves in greater detail than the main paper.
\subsection{QCQP Formulation}
Throughout this section we slightly abuse our notation by using $e$ to refer both to a directed edge $e = (i, j) \in \EqualityEdges$ as well as an integer ($e \in  \Indices{\Cardinality{\EqualityEdges}}$) corresponding to a fixed index of this same edge.
The incidence matrix is
\begin{equation}
	\IncidenceMatrix{\EqualityEdges}_{i, e} = 
	\begin{cases}
	\hspace{\minuslength} 1 \enspace \mathrm{if} \enspace e \in \EnteringEdges{i}, \\
	-1 \enspace \mathrm{if} \enspace e \in \LeavingEdges{i}, \\
	\hspace{\minuslength} 0 \enspace \mathrm{otherwise},
	\end{cases}
\end{equation}
where $\LeavingEdges{i}$ and $\EnteringEdges{i}$ are the set of edges leaving and entering $i \in \Vertices$, respectively. 
Thus, the columns 
\begin{equation}
	\Matrix{P}\IncidenceMatrix{\EqualityEdges}^{(e)} = \Matrix{P}^{(j)} - \Matrix{P}^{(i)}
\end{equation}
each refer to relative position of vertices $i$ and $j$.
The diagonal elements of the product
\begin{equation}
	\IncidenceMatrix{\EqualityEdges}^\T\Matrix{P}^\T \Matrix{P} \IncidenceMatrix{\EqualityEdges} \in \Real^{\Cardinality{\EqualityEdges} \times \Cardinality{\EqualityEdges}}
\end{equation} 
are therefore equal to
\begin{equation}
	\Distance(e) = \Norm{\Matrix{P}^{(j)} - \Matrix{P}^{(i)}}^2, \enspace e = (i, j) \in \EqualityEdges.
\end{equation}
Recalling that $\Vector{d}_e = \Distance(e)^2$ leads us to the compact expression for our equality constraints found in the main paper:
\begin{equation} \label{eq:equality_constraints}
	\diag{\IncidenceMatrix{\EqualityEdges}^\T \Matrix{P}^\T \Matrix{P} \IncidenceMatrix{\EqualityEdges}} = \Vector{\ell}.
\end{equation}

\subsection{SDP Constraints}
After deriving the feasibility QCQP formulation of IK, we apply an SDP relaxation. 
The lifted matrix variable
\begin{equation} \label{eq:lifted_Z_variable}
	\Matrix{Z}(\Matrix{X}) = [\Matrix{X} \enspace \Identity_\Dim]^\T [\Matrix{X} \enspace \Identity_\Dim] = 
	\begin{bmatrix}
	  \Matrix{X}^\T \Matrix{X} & \Matrix{X}^\T \\
	  \Matrix{X} & \Identity_\Dim	
	\end{bmatrix}
\in \PSDMatrices{2n+\Dim} 
\end{equation}
allows us to write a number of degree-2 or lower polynomial expressions in $\Matrix{X}$ as linear expressions of $\Matrix{Z}$. 
Since $\Matrix{Z}_{i,j} = \Vector{x}_i^\T \Vector{x}_j \ \forall i, j \leq 2n$, each equation in \ref{eq:equality_constraints} that is between two variables $\Vector{x}_k, \Vector{x}_l$, $k \neq l$ can be written $\Trace{\Matrix{A}\Matrix{Z}} = \Distance((k, l))$, where
\begin{equation}
	\Matrix{A}_{j,i} = \Matrix{A}_{i,j} = 
	\begin{cases}
	\hspace{\minuslength} 1 \enspace \mathrm{if} \enspace i = j \in \{k, l\}, \\
	-1 \enspace \mathrm{if} \enspace i = k,\ j=l, \\
	\hspace{\minuslength} 0 \enspace \mathrm{otherwise}.
	\end{cases}
\end{equation}
Similarly, the expression for the squared distance between a variable $\Vector{x}_k$ and some anchor $\Anchor$ (or obstacle centre $\Vector{c}$) can be encoded with matrix $\Matrix{M} \in \SymmetricMatrices{2n+\Dim}$ where
\begin{align}
	\Matrix{M}_{k, k} &= 1, \notag \\
	\Matrix{M}_{2n:2n+d, k} &= -\Anchor, \\
	\Matrix{M}_{k, 2n:2n+d} &= -\Anchor^\T, \notag\\
	\Matrix{M}_{2n+d, 2n+d} &= \Norm{\Anchor}^2. \notag
\end{align}
Likewise, linear constraints can be enforced with matrices manipulating the elements in $\Matrix{Z}_{2n:2n+d, 1:2n}$ and its symmetric counterpart. 
These constraints can be collected in the linear maps 
\begin{align}
	\LinearMap(\Matrix{Z}) &= \Vector{a}, \enspace \LinearMap: \SymmetricMatrices{2n+\Dim} \rightarrow \Real^{m+\Dim^2 + \Cardinality{\Vertices_{\mathrm{p}}}}, \\
	\LinearInequalityMap(\Matrix{Z}) &\leq \Vector{b}, \enspace \LinearInequalityMap: \SymmetricMatrices{2n+\Dim} \rightarrow \Real^{\Cardinality{\Obstacles}},
\end{align}
which appear in our SDP problem formulation.
The final constraint worth noting, which contributes $\Dim^2$ to the dimension of the codomain of $\mathcal{A}$, simply arises from the requirement that the bottom-right $\Dim \times \Dim$ diagonal of $\Matrix{Z}$ is equal to $\Identity_{\Dim}$. 

\subsection{Geometric Interpretation}
This section provides details on the toy SDP formulation in Section IV-C of the main paper.
Recall the QCQP formulation for the IK problem of reaching a point $\Vector{w} \in \Real^2$ with the end-effector of a simple planar 2-DOF manipulator:
\begin{align} \label{eq:toy_qcqp}
\mathrm{find}  &\quad \Vector{x} \in \Real^2 \notag \\
\text{\emph{s.t.}}  &\quad \Norm{\Vector{x}}^2 = 1, \\
&\quad \Norm{\Vector{x} - \Vector{w}}^2 = 1, \notag \\
&\quad \Norm{\Vector{x}- \Vector{o}}^2 \geq 0.25, \notag	
\end{align}
where $\Vector{o} = [1,\ 0]^\T$ is the position of a unit-diameter circular obstacle.
For $\Vector{w} = [1,\ 1]^\T$, the insets of Figure 3 in the main paper show that of the two candidate solutions to this problem, the ``elbow down" configuration in the bottom right collides with the obstacle at $\Vector{o}$ (partially depicted as a blue semicircle). 
Homogenizing (\ref{eq:toy_qcqp}) with $s^2 = 1$ and lifting to the rank-1 matrix variable
\begin{equation}
	\Matrix{Z}(\Vector{x}) = \bbm \Vector{x} \\ s \ebm \bbm \Vector{x}^\T & s \ebm
\end{equation}
lets us apply the SDP relaxation $\Matrix{Z} \succeq \Matrix{0}$ to yield:
\begin{align} \label{eq:toy_sdp}
\mathrm{find} &\quad \Matrix{Z} \in \PSDMatrices{3} \notag \\
\text{\emph{s.t.}} &\quad \Trace{\Matrix{A}_0\Matrix{Z}} = 1, \\
&\quad \Trace{\Matrix{A}_1\Matrix{Z}} = 1, \notag \\
&\quad \Trace{\Matrix{A}_2\Matrix{Z}} = 1, \notag \\
&\quad \Trace{\Matrix{A}_3\Matrix{Z}} \geq 0.25, \notag	
\end{align}
where 
\begin{equation}
	\Matrix{A}_0 = \begin{bmatrix}
		1 & 0 & 0\\
		0 & 1 & 0\\
		0 & 0 & 0
	\end{bmatrix}, \quad
	\Matrix{A}_1 = \begin{bmatrix}
		1 & 0 & -1\\
		0 & 1 & -1\\
		-1 & -1 & 2
	\end{bmatrix}
\end{equation}
describe the unit link length constraints and
\begin{equation}
	\Matrix{A}_2 = \begin{bmatrix}
		0 & 0 & 0\\
		0 & 0 & 0\\
		0 & 0 & 1
	\end{bmatrix}, \quad
	\Matrix{A}_3 = \begin{bmatrix}
		1 & 0 & -1\\
		0 & 1 & 0\\
		-1 & 0 & 1
	\end{bmatrix}
\end{equation}
describe the homogenization equation ($s^2 = 1$) and the obstacle avoidance constraint, respectively.

\subsection{Link and Self-Collision Avoidance}
In order to implement collision avoidance on parts of the robot that are not represented by the points in $\Matrix{X}$, we can introduce any number of auxiliary variables $\Vector{y} \in \Real^\Dim$ that are fixed between two points $\Vector{x}_i$ and $\Vector{x}_j$ for some $(i,j) \in \Edges_{\text{eq}}$.  
We can parameterize the interior of the line segment connecting $\Vector{x}_i$ and $\Vector{x}_j$ with $\alpha \in (0, 1)$ and constrain $\Vector{y}$ to lie at some point of our choosing on this line segment:
\begin{equation} \label{eq:auxiliary_variable}
	\Vector{y} = (1-\alpha)\Vector{x}_i + \alpha\Vector{x}_j.
\end{equation}
This auxiliary point can be used in collision avoidance constraints between $\Vector{y}$ and obstacles:
\begin{equation}
	\Norm{\Vector{y} - \Vector{c}_k}^2 \geq l_k^2 \enspace \forall k \in \Obstacles.
\end{equation}

Additionally, \Ours can easily incorporate self-collision constraints with distance inequalities between variables:
\begin{equation}
	\Norm{\Vector{x}_i - \Vector{x}_j}^2 \geq \epsilon_{i,j} \enspace \forall (i,j) \in \Edges_{\text{eq}},
\end{equation}
where $\epsilon_{i,j}$ is any user-defined threshold, ideally based on robot geometry. 
Equations of this form can also replace $\Vector{x}_i$ or $\Vector{x}_j$ with auxiliary variables defined in \Cref{eq:auxiliary_variable}.
Since the constraints defined in this section are either linear or distance-geometric, they are supported by the ``rank-$\Dim$" semidefinite relaxation of \Cref{eq:lifted_Z_variable}.

\section{Additional Experiments} 
This section summarizes two small experiments not covered in the main paper. 

\subsection{Infeasibility Certification}
Convex relaxations can often be used certify the infeasibility of a problem.
To determine whether \Ours is capable of infeasibility certification, we conducted a simple experiment. 
Using the \textit{Cube} environment and a closed-form analytic solver for the UR10, we gave \Ours 10,000 goal poses with the following properties:
\begin{enumerate}
	\item the analytic UR10 solver deemed the goal pose \emph{infeasible} by checking for collisions over all 8 possible solutions,
	\item the end-effector was not in collision with an obstacle, and
	\item the point attached to the end-effector's parent joint was not in collision with an obstacle. 
\end{enumerate} 
Goal poses that are infeasible because of conditions 2) and 3) are easy for \Ours to certify as infeasible because the Euclidean distance constraints they violate are unavoidable, even in higher-rank solutions representing configurations embedded in $\Real^{\Dim'}$ where $\Dim' > \Dim$.
Unfortunately, the infeasible cases that remain are much more challenging, with \Ours certifying a mere 152 out 10,000 (1.52\%) as infeasible. 

\subsection{Nuclear Norm Heuristic}
\begin{figure}
\centering 
\includegraphics[width=\columnwidth]{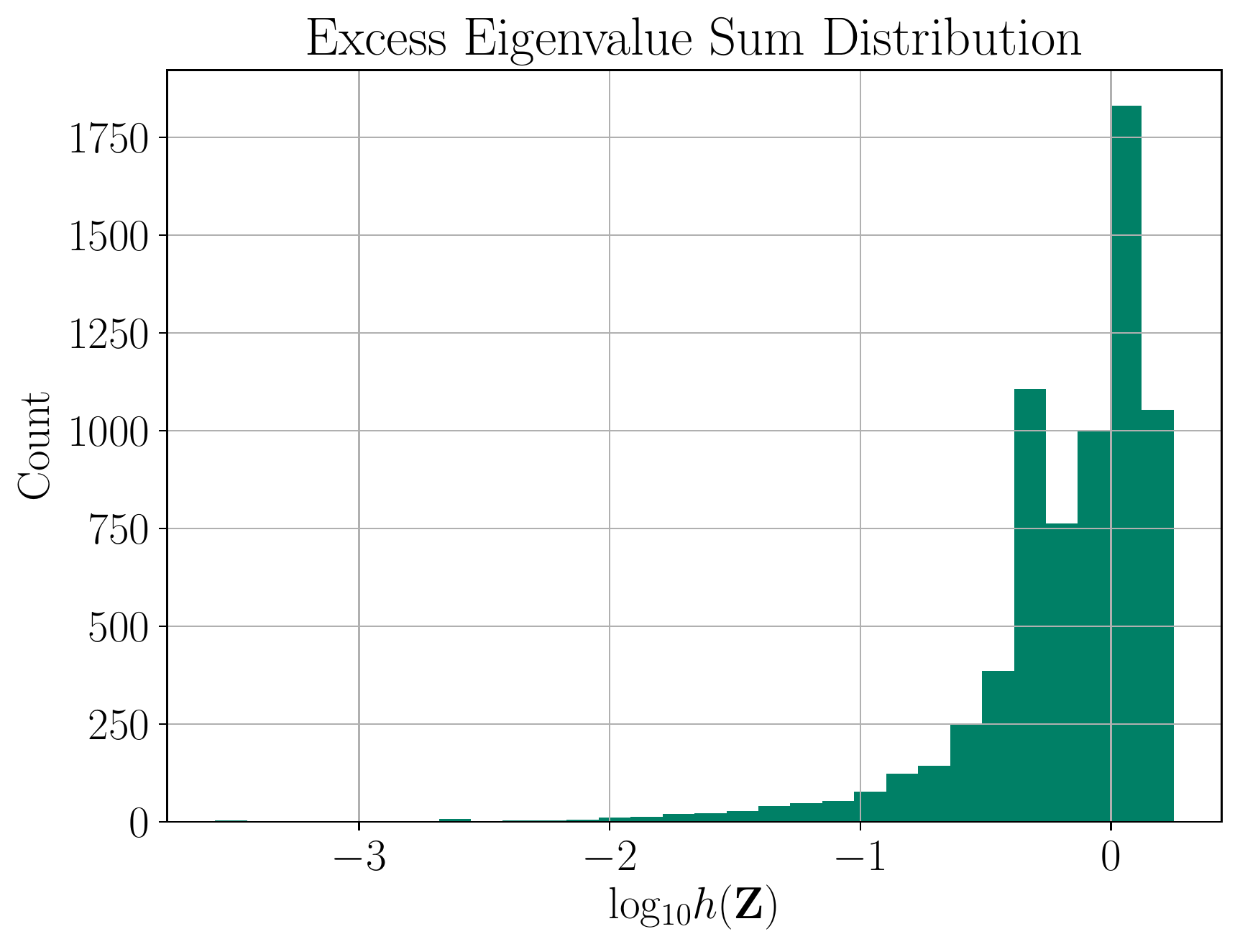}
\caption{Distribution of $\log_{10}(h(\Matrix{Z})$ over 7,000 runs of our SDP formulation with the nuclear norm heuristic ($\Matrix{C} = \Matrix{I}$).}
\vspace{-0.6cm}
\label{fig:nuclear_norm_histogram}
\end{figure}
The first iteration of \Ours uses cost matrix $\Matrix{C} = \Identity$, which is equivalent to the popular nuclear norm heuristic.
The nuclear norm is an effective surrogate for low-rank matrix recovery in other problems, but in this section we demonstrate that it is inappropriate for IK.
\Cref{fig:nuclear_norm_histogram} shows the distribution of $\log_{10}(h(\Matrix{Z}))$ over 7,000 experiments with feasible UR10 goal poses in an obstacle-free environment, where
\begin{equation}
	h(\Matrix{Z}) = \sum_{i=\Dim +1}^{2n+\Dim} \lambda_{i}(\Matrix{Z})
\end{equation}
is our ``excess rank" heuristic.
The convergence threshold for $h(\Matrix{Z})$ we used for our experiments with \Ours in the main paper was 10$^{-6}$, with some high quality solutions recovered for $h(\Matrix{Z}) \approx 10^{-3}$. 
As you can see in \Cref{fig:nuclear_norm_histogram}, the nuclear norm heuristic led to very few instances with $h(\Matrix{Z}) < 10^{-2}$ and is therefore only appropriate as an initialization for \Ours, which typically required fewer than 10 iterations to converge below our threshold of 10$^{-6}$.

\subsection{SDP Solver Performance}
\begin{figure}
\centering 
\includegraphics[width=\columnwidth]{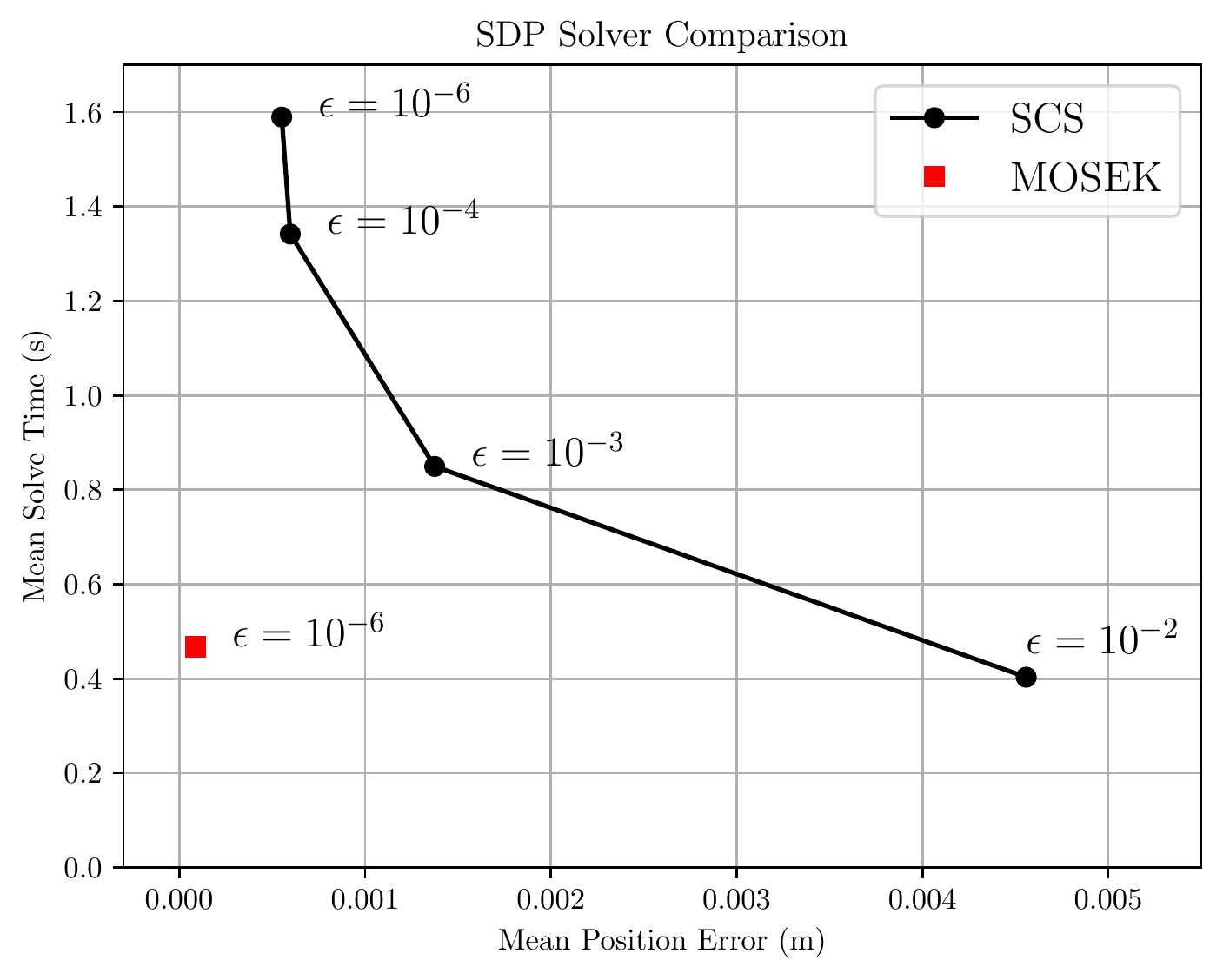}
\caption{A comparison of MOSEK (used in this paper's experiments) with SCS for multiple values of $\epsilon$, the convergence tolerance. The reported mean solve times and position errors are over 100 randomly generated IK problems. Each algorithm was allowed up to 1,000 iterations. }
\label{fig:sdp_solver_comparison}
\end{figure}
This section provides a comparison of two SDP solvers: SCS, a conic operator splitting method~\cite{odonoghue_scs_2016}, and MOSEK, an interior point optimizer~\cite{andersen_mosek_2000}.
These solvers were selected because they use different approaches, and because of their interfaces with the \texttt{cvxpy} modelling language~\cite{diamond_cvxpy_2016} used to implement \Ours.
\Cref{fig:sdp_solver_comparison} demonstrates that MOSEK is the far superior option on a dataset of 100 randomly generated IK problems using the 9-DOF manipulator in the \textit{table} environment. 
In spite of its use of warm starting, the performance of SCS is dominated by MOSEK. 
SCS is only able to run faster than MOSEK when it tolerates end-effector position errors an order of magnitude greater than MOSEK's.

Finally, in order to highlight the importance of choosing a fast SDP solver, we measured the total runtime of the closed-form eigendecomposition solution when solving the 100 IK problem instances with MOSEK.
The mean runtime of this procedure is 0.56 ms, with a standard deviation of 0.26 ms.
This is dwarfed by the nearly half-second runtime of MOSEK on the main SDP.

\section{Limitations and Future Work}
This section describes a number of promising research directions that can address some of the shortcomings and limitations of \Ours. 

\subsection{Rank-1 Relaxation and Ellipsoids}
Like SNL, our IK problem leads to a rank-$\Dim$ SDP relaxation because of the $\Dim$-dimensional nature of our spatial problem.
More concretely, we saw in the previous section that since we are only constraining the squared \emph{distances} between points in $\Real^{\Dim}$, quadratic terms in these constraints appear as inner products between variables (e.g., $\Vector{x}_i^\T \Vector{x}_j$).
In general, QCQPs do not contain this structure, and the lifted $\Matrix{Z}$ variable would be a ``rank-1" lifting.
For our problem, a rank-1 SDP relaxation, like the one used for the \emph{pierogi} example would take the form
\begin{equation}
	\Matrix{Z} = [\ColVec{\Matrix{X}}^\T \ 1]^\T [\ColVec{\Matrix{X}}^\T \ 1] \in \PSDMatrices{2\Dim n + 1}, 
\end{equation}
where $\ColVec{\cdot}: \Real^{a \times b} \rightarrow \Real^{ab}$ is a column-wise vectorization of its matrix argument (i.e., it ``vertically" stacks $b$ columns).
As you can see, this $\Matrix{Z}$ variable is much larger, leading to greater computation and memory requirements in SDP solvers, and our rank-$\Dim$ relaxation, by contrast, exploits the spatial structure of our problem.
However, since the generic rank-1 relaxation allows us to work with any quadratic constraint, in future work we are interested in exploring its ability to model constraints like arbitrary ellipsoidal obstacles:
\begin{equation}
	(\Vector{x}_i - \Vector{c})^\T \Matrix{A} (\Vector{x}_i - \Vector{c}) \geq l^2.
\end{equation}

\subsection{Chordal Sparsity}
A graph is \emph{chordal} if every cycle of length four or greater has a chord (an edge that is not part of the cycle but connects two vertices in the cycle).
In our problem, the edges of $\Graph$ describe points whose distances are constrained. 
Since the SDP variable $\Matrix{Z}$ contains the dot product of point variables (i.e., $Z_{ij} = \Matrix{x}_i^\T \Matrix{x}_j$), $\Graph$ also describes the elements of $\Matrix{Z}$ that appear together in constraints. 
In this section, we demonstrate that $\Graph$ is chordal due to a sparsity pattern induced by the fact that distance constraints only affect points fixed to neighbouring joints. 
This ``chordal sparsity" can be exploited to speed up the solution of large SDPs with the methods reviewed in \cite{majumdar_survey_2019}.
In future work, we intend to use these methods within \Ours to quickly solve IK problems for tree-like robots with many redundant degrees of freedom.

Assume that $\Graph = (\Vertices, \EqualityEdges, \Distance)$ is the distance constraint graph for a tree-like robot (i.e., no loops in its joints). 
Since the robot's joints and links describe a tree, the only cycles in $\Graph$ occur in the cliques described by $\Vector{p}_i$, $\Vector{q}_i$, $\Vector{p}_j$, and $\Vector{q}_j$ for neighbouring joints $i$ and $j$. 
Since the cycles occur within cliques, they are chordal (because cliques are complete by definition). 
In order to apply the methods in \cite{majumdar_survey_2019}, we would additionally need to ensure that the workspace constraints and cost function matrix $\Matrix{C}$ also exhibit this sparsity pattern.
Spherical and planar obstacle constraints clearly do not ruin the chordal sparsity, since each constraint only involves a single point $\Vector{p}_i$. 
Finally, the work on optimal power flow in \cite{wang_chordal_2018} describes a heuristic method for constructing $\Matrix{C}$ in a manner that preserves problem sparsity.
The COSMO algorithm \cite{garstka_cosmo_2019} has a freely available implementation that is able to automatically exploit chordal sparsity in conic optimization problems. 

\subsection{Joint Angle Limits}
Quadratic joint angle limits for revolute manipulators can be defined in a manner similar to the method used in \cite{maricInverseKinematicsSerial2020} on planar and spherical chains. 
Unfortunately, without the addition of auxiliary variables, joint angle ranges which are not symmetric about the ``straightened" configuration are impossible to enforce. 
Crucially, this precludes the use of joints that behave like fingers or knees (i.e., only able to bend in one direction away from the straightened configuration) in IK problems we wish to solve with \Ours.
This need for ``directionality" in angular constraints can be addressed with the addition of constraints involving the cross product of differences between adjacent points, which we leave for future work. 

\subsection{``Tetrahedral" Robot Link Geometry}
The work in this paper assumes that consecutive joints have axes of rotation that are coplanar (i.e., either parallel or intersecting).
This requirement ensures that for consecutive joints $i$ and $j$, the points $\Vector{p}_i, \Vector{q_i}, \Vector{p}_j$, and $\Vector{q}_j$ form a quadrilateral or triangle.
When the axes are \emph{not} coplanar, these points form a tetrahedron that we are specifying from distances alone. 
Generic tetrahedra are chiral in that their reflections cannot be reproduced by rigid transformations. 
Therefore, our method of enforcing a tetrahedral joint pair's structure with distances alone allows \Ours to search a feasible set that contains physically unrealizable configurations.
Much like asymmetrical joint limits, this problem can be addressed with the addition of auxiliary variables and constraints involving cross-products, but the effect of these additions on the accuracy and performance of \Ours remains to be seen. 

\bibliographystyle{IEEEtran}
\bibliography{robotics_abbrv,references}